%% file: main_tmp.tex
\documentclass{article}

\usepackage[eandd,preprint]{neurips_2026}

\usepackage{graphicx}
\usepackage{amsmath}
\usepackage{float}
\usepackage[linesnumbered,ruled,algo2e]{algorithm2e}%

\usepackage[utf8]{inputenc} 
\usepackage[T1]{fontenc}    
\usepackage{hyperref}       
\usepackage{url}            
\usepackage{booktabs}
\usepackage{cellspace}
\usepackage{amsfonts}       
\usepackage{nicefrac}       
\usepackage{microtype}      
\usepackage{xcolor}         
\usepackage{wrapfig}        
\usepackage{enumitem}
\usepackage{etoc}

\newcommand{\kc}[1]{\textcolor{teal}{{(KC) #1}}}

\newcommand{\TSc}[1]{\textcolor{orange}{{(TS) #1}}}

\usepackage[colorinlistoftodos,prependcaption,textsize=tiny]{todonotes}
\providecommand{\tdotoggle}{1}
 \ifnum\tdotoggle=1
 \fi

\title{N$^2$-Bench: A Benchmark Suite for Scalar and Distributional Matrix Completion across Real-World Domains}

\author{%
    Caleb Chin \\
    Cornell University \\
    \And
    Aashish Khubchandani \\
    Cornell University \\
    \And
    Harshvardhan Maskara \\
    Cornell University \\
    \And
    Kyuseong Choi \\
    Cornell University \\
    \And
    Jacob Feitelberg \\
    Columbia University \\
    \And
    Albert Gong \\
    Cornell University \\
    \And
    Manit Paul  \\
    University of Pennsylvania
    \And
    Tathagata Sadhukhan \\
    Cornell University
    \And
    Dwaipayan Saha \\
    Columbia University
    \And
    Anish Agarwal \\
    Columbia University
    \And
    Raaz Dwivedi \\
    Cornell University
}

\input{cref_setup-2}
\input{macros}

\newcommand{\dist}[1]{\mu_{#1}}
\newcommand{\sample}[1]{X_{#1}}
\newcommand{\measurement}[1]{Z_{#1}}
\newcommand{\missing}[1]{A_{#1}}
\newcommand{\meanparam}[1]{\theta_{#1}}

\newcommand{\nn}{\mathsf{NN}}
\newcommand{\rownn}{\mathsf{RowNN}}
\newcommand{\colnn}{\mathsf{ColNN}}
\newcommand{\tsnn}{\mathsf{TSNN}}
\newcommand{\drnn}{\mathsf{DRNN}}
\newcommand{\awnn}{\mathsf{AWNN}}

\newcommand{\kernelnn}{\mathsf{KernelNN}}
\newcommand{\wassersteinnn}{\mathsf{W_2NN}}

\newcommand{\dr}{\mathsf{DR}}
\newcommand{\ts}{\mathsf{TS}}
\newcommand{\lib}{$\mathbf N^2$}

\newcommand{\tuning}{\eta}
\newcommand{\distance}[1]{\rho_{#1}}
\newcommand{\neighbors}[1]{\mbf{N}_{#1}}

\newcommand{\mmd}{\mathsf{MMD}}
\newcommand{\wasserstein}{\mathsf{W}_2}
\newcommand{\hatwasserstein}{\what{\mathsf{W}}_2}

\newcommand{\distancemod}{\textsc{Distance}}
\newcommand{\averagemod}{\textsc{Average}}
\newcommand{\composemod}{\textsc{Compose}}

\newcommand{\subtractmod}{\textsc{Subtract}}
\newcommand{\entrydist}[1]{\varphi(#1)}
\newcommand{\entrydistest}[1]{\what{\varphi}(#1)}

\newcommand{\weight}[1]{w_{#1}}

\newcommand{\syntheticmeasurement}[1]{\tilde{Z}_{#1}}

\begin{document}

\maketitle

\begin{abstract}

Matrix completion methods are often tested on synthetic low-rank matrices or a small set of recommender-system datasets, so it remains unclear how they perform when missingness is structured, adaptive, or confounded. We introduce N$^2$-Bench, a benchmark suite and open-source evaluation harness for matrix completion on realistic data. The suite includes tasks from mobile health, recommender systems, causal panel data, and LLM evaluation, and supports both scalar matrix completion and distributional matrix completion, where an entry may be an empirical distribution rather than a single value. Each task includes data loaders, train-validation-test splits, metrics, and a common interface for comparing nearest-neighbor, spectral, optimization-based, and causal inference baselines.
We also release N$^2$, a modular Python library for nearest-neighbor matrix completion, covering row-wise, column-wise, two-sided, doubly robust, adaptively weighted, and distributional variants. Using N$^2$-Bench, we find that conclusions from idealized synthetic experiments can be misleading: methods that work well under simple random missingness often struggle on real data, while nearest-neighbor methods are often competitive and more stable across domains.

\end{abstract}

\input{neurips2026_tmp/intro}
\input{neurips2026_tmp/benchmark-datasets}
\input{neurips2026_tmp/case-study}
\input{neurips2026_tmp/llm_evaluation_new}
\input{neurips2026_tmp/library}
\input{neurips2026_tmp/conclusion}


\bibliographystyle{alpha}
\bibliography{ref}


\onecolumn
\appendix

\etocsettocstyle{}{}
    \etocdepthtag.toc{mtappendix}
    \etocsettagdepth{mtchapter}{none}
    \etocsettagdepth{mtappendix}{section}
    \etocsettagdepth{mtappendix}{subsection}
    \etocsettagdepth{mtappendix}{subsubsection}
    {\tableofcontents}

\input{neurips2026_tmp/nn-overview}

\input{neurips2026_tmp/nn-new}

\input{appendices/guarantees}

\input{appendices/nn-algo}

\input{appendices/cv}
\input{appendices/case_study_details}

\newpage

\end{document}

%% file: cref_setup-2.tex
\usepackage{thmtools,thm-restate}
\usepackage[capitalize]{cleveref}  

\usepackage{autonum}

\newtheorem{lemma}{Lemma}
\newtheorem{definition}{Definition}
\newtheorem{example}{Example}
\newtheorem{remark}{Remark}

\newtheorem{assumption}{Assumption}
\newtheorem{corollary}{Corollary}
\newtheorem{proposition}{Proposition}

\newtheorem{myexample}{Example}

\newtheorem{myproposition}{Proposition}

\newtheorem{mycorollary}{Corollary}

\newtheorem{mylemma}{Lemma}

\newtheorem{mydefinition}{Definition}

\newtheorem{myremark}{Remark}

\newtheorem{myassumption}{Assumption}

\crefname{appendix}{App.}{Apps.}
\crefname{subsubsubappendix}{App.}{Apps.}
\crefname{equation}{}{}
\crefname{lemma}{Lem.}{Lems.}
\crefname{theorem}{Thm.}{Thms.}
\Crefname{theorem}{THM.}{THMS.}
\crefname{Corollary}{Cor.}{Cors.}
\crefname{algorithm}{Alg.}{Algs.}

\crefname{section}{Sec.}{Secs.}
\crefname{table}{Tab.}{Tabs.}
\crefname{remark}{Rem.}{Rems.}
\crefname{definition}{Def.}{Defs.}
\crefname{Proposition}{Prop.}{Props.}
\crefname{myremark}{Rem.}{Rems.}
\crefname{mylemma}{Lem.}{Lems.}
\Crefname{mylemma}{LEM.}{LEMS.}
\crefname{mydefinition}{Def.}{Defs.}
\crefname{myproposition}{Prop.}{Props.}
\Crefname{myproposition}{PROP.}{PROPS.}
\crefname{mycorollary}{Cor.}{Cors.}
\Crefname{mycorollary}{COR.}{CORS.}
\crefname{myassumption}{Assum.}{Assums.}
\crefname{figure}{Fig.}{Figs.}
\crefname{myexample}{Ex.}{Exs.}
\crefname{enumi}{}{}
\crefname{name}{}{} 

%% file: macros.tex
\usepackage{amsmath}

\newcommand{\rowimputelink}{\hyperref[algo:row_imputer]{\textsc{Row-imputer}}\xspace}

\newcommand{\wgtadjlink}{\hyperref[algo:wgt_adjuster]{\textsc{Weight-Adjuster}}\xspace}

\newcommand{\drow}[1][i']{\widehat{\rho}^2\parenth{#1,i}}

\newcommand{\drowmeanfrac}[1][\mathcal{A}_j]{\frac{1}{\abss{#1}}\sum_{i\in#1}\drow}

\newcommand{\Wgt}[1][j]{\what{w}^{(i_0)}_{i,#1}}
\newcommand{\pseqann}[1][n]{(a_i)_{i=1}^{#1}} 

\newcommand{\pseqnn}[2]{(#1_i)_{i=1}^{#2}}

\newcommand{\col}{\mrm{col}}
\newcommand{\row}{\mrm{row}}

\newcommand{\indicator}{\mbf 1}

\newcommand{\sumn}[1][i]{\sum_{#1=1}^n}

\newcommand{\x}{x}

\newcommand{\axi}[1][i]{\x_{#1}}

\newcommand{\dirac}{\mbi{\delta}}

\newcommand{\kernel}{\mbf{k}}

\newcommand{\eps}{\epsilon}

\newcommand{\ol}[1]{\overline{#1}}

\newcommand{\pseqxn}[1][n]{(\axi[i])_{i\geq 1}} 
\newcommand{\pseqxnn}[1][n]{(\axi[i])_{i=1}^n} 

\newcommand{\brackets}[1]{\left[ #1 \right]}

\newcommand{\parenth}[1]{\left( #1 \right)}

\newcommand{\sbraces}[1]{\{ #1  \}}
\newcommand{\braces}[1]{\left\{ #1 \right \}}

\newcommand{\abss}[1]{\left| #1 \right |}

\newcommand{\real}{\ensuremath{\mathbb{R}}}

\newcommand{\Prob}{\ensuremath{{\mathbb{P}}}}

\def\balign#1\ealign{\begin{align}#1\end{align}}
\def\baligns#1\ealigns{\begin{align*}#1\end{align*}}
\def\balignat#1\ealign{\begin{alignat}#1\end{alignat}}
\def\balignats#1\ealigns{\begin{alignat*}#1\end{alignat*}}
\def\bitemize#1\eitemize{\begin{itemize}#1\end{itemize}}
\def\benumerate#1\eenumerate{\begin{enumerate}#1\end{enumerate}}

\newenvironment{talign*}
 {\csname align*\endcsname}
 {\endalign}
\newenvironment{talign}
 {\csname align\endcsname}
 {\endalign}

\def\balignst#1\ealignst{\begin{talign*}#1\end{talign*}}
\def\balignt#1\ealignt{\begin{talign}#1\end{talign}}

\newcommand{\qtext}[1]{\quad\text{#1}\quad}

\let\originalleft\left
\let\originalright\right
\renewcommand{\left}{\mathopen{}\mathclose\bgroup\originalleft}
\renewcommand{\right}{\aftergroup\egroup\originalright}

\def\tinycitep*#1{{\tiny\citep*{#1}}}
\def\tinycitealt*#1{{\tiny\citealt*{#1}}}
\def\tinycite*#1{{\tiny\cite*{#1}}}
\def\smallcitep*#1{{\scriptsize\citep*{#1}}}
\def\smallcitealt*#1{{\scriptsize\citealt*{#1}}}
\def\smallcite*#1{{\scriptsize\cite*{#1}}}

\def\mbi#1{\boldsymbol{#1}} 
\def\mbf#1{\mathbf{#1}}

\def\mc#1{\mathcal{#1}}
\def\mrm#1{\mathrm{#1}}

\def\tbf#1{\textbf{#1}}

\def\<{\left\langle} 
\def\>{\right\rangle}

\def\what#1{\widehat{#1}}

\def\indic#1{\indicator({#1})} 

\def\Var{\mrm{Var}} 
\def\Vararg#1{\Var\left[{#1}\right]}

\providecommand{\argmin}{\mathop\mathrm{arg min}}

\ifdefined\nonewproofenvironments\else
\ifdefined\ispres\else

\newenvironment{proof-sketch}{\noindent\textbf{Proof Sketch}
  \hspace*{1em}}{\qed\bigskip\\}
\newenvironment{proof-idea}{\noindent\textbf{Proof Idea}
  \hspace*{1em}}{\qed\bigskip\\}
\newenvironment{proof-of-lemma}[1][{}]{\noindent\textbf{Proof of Lemma {#1}}
  \hspace*{1em}}{\qed\\}
\newenvironment{proof-of-theorem}[1][{}]{\noindent\textbf{Proof of Theorem {#1}}
  \hspace*{1em}}{\qed\\}
\newenvironment{proof-attempt}{\noindent\textbf{Proof Attempt}
  \hspace*{1em}}{\qed\bigskip\\}



%% file: neurips2026_tmp/intro.tex
\section{Introduction}

Matrix completion is a well-established field that supplies practitioners with many tools to recover underlying matrices using partial or even noisy observations~\cite{hastie2015matrix,chatterjee2015matrix,keshavan2010matrix}, with applications spanning recommendation systems~\cite{koren2009matrix,recht2011simpler}, causal panel data~\cite{bai2009panel,athey2021matrix,agarwal2023causal}, mobile health~\cite{klasnja2019efficacy}, and LLM evaluation. Yet, most methods are benchmarked on synthetic low-rank matrices or a narrow set of recommender-system datasets under random missingness, leaving it unclear how they perform when missingness is structured, adaptive, or confounded. Meanwhile, nearest neighbor~(NN) methods have emerged as effective alternatives that target individual entries via matching and are robust to confounded observation patterns~\cite{li2019nearest,ma2019missing,dwivedi2022counterfactual,agarwal2023causal}, including recent extensions to distributional settings~\cite{feitelberg2024distributional,choi2024learning}. However, there is no unified benchmark or package
\footnote{\url{https://github.com/aashish-khub/NearestNeighbors}}
that lets users systematically compare these methods across realistic tasks.

\newcommand{\starnn}{\mathsf{StarNN}}
\newcommand{\autonn}{\mathsf{AutoNN}}

\paragraph{Our contributions.}
We make the following contributions.
(1)~We introduce \lib{}-\textbf{Bench}, a benchmark suite for matrix completion on realistic data. It spans four domains---mobile health, recommender systems, causal panel data, and LLM evaluation---and supports both scalar and distributional matrix completion. Each task ships with data loaders, train-validation-test splits, metrics, and a common interface for comparing nearest-neighbor, spectral, optimization-based, and causal inference baselines.
(2)~Using \lib{}-Bench, we provide the first comprehensive comparison of NN methods on real-world matrix completion problems and find that methods excelling under random missingness often degrade on real data, while NN methods are consistently competitive and more stable across domains.
(3)~We release \lib{}, a modular Python library for NN-based matrix completion covering row-wise, column-wise, two-sided, doubly robust, adaptively weighted, and distributional variants.
(4)~We introduce $\autonn$, a new NN variant that automatically interpolates between debiasing and denoising, and demonstrate the ease of developing new methods within our framework.

\subsection{Related work}
We contextualize our contributions in the context of both nearest neighbors as a general algorithm and matrix completion specifically.

\paragraph{Nearest neighbors.}

As introduced above, nearest neighbor methods are widely used for non-parametric regression, classification, and pattern recognition~\cite{cover1967nearest,cannings2020local}. Recently, nearest neighbor methods were introduced as an effective algorithm for matrix completion problems~\cite{li2019nearest,agarwal2023causal,dwivedi2022counterfactual}, especially when the missingness depended on observations and unobserved confounders. The fact that nearest neighbor methods target a single entry at a time via matching makes them effective against various types of missing patterns. The class of algorithms has grown to account for a wide range of applications involving scalar or distributional settings; for instance, nearest neighbors are used for evaluating the personalized recommendation effect of a healthcare app on the average and the distribution of physical step counts~\cite{dwivedi2022counterfactual,choi2024learning,feitelberg2024distributional,sadhukhan2024adaptivity} and are also used for policy evaluations~\cite{agarwal2023causal}.
Our library, \lib{}, is designed to easily implement the vanilla scalar versions of nearest neighbors~\cite{li2019nearest,dwivedi2022counterfactual} as well as its unweighted~\cite{dwivedi2022doubly,sadhukhan2024adaptivity} and weighted~\cite{sadhukhan2025adaptivelyweightednearestneighborsmatrix} variants. Finally, \lib{} is also capable of distributional matrix completion~\cite{choi2024learning,feitelberg2024distributional}.

\paragraph{Other matrix completion methods.}


Universal singular value thresholding (USVT), proposed in \cite{chatterjee2015matrix,bhattacharya2022matrix}, is a classical spectral-based method for performing matrix completion; its core functionality is based on a singular value decomposition of the matrix and thresholding the singular values.
SoftImpute, introduced by \cite{hastie2015matrix}, is another widely used optimization-based algorithm for matrix completion. The algorithm computes ridge-regression updates of the low-rank factors iteratively and finally soft-thresholds the singular values to impose a nuclear norm penalty. 
Notably USVT and SoftImpute have provable guarantees when missingness is completely at random, but empirically fail when the missing pattern depends on the observed entries or the unobserved confounders~\cite{agarwal2023causal}. Our real-world analysis in \cref{sec:case-study} once again demonstrates this point.



\paragraph{Existing software for matrix completion and nearest neighbors.}

Scikit-Learn \cite{scikit-learn}, a popular Python package for machine learning tools, implements a simple k-nearest neighbor algorithm for imputing missing values in a feature matrix. However, their implementation is designed for the feature matrix setting. So, neighbors are only defined across samples (row-wise). Additionally, they do not provide any implementation for more advanced nearest neighbor algorithms, nor does their package allow for easy extendability like our proposed package. 


%% file: neurips2026_tmp/benchmark-datasets.tex
\section{Benchmark: Datasets and Methods}\label{sec:benchmark}

We now describe the datasets comprising \lib{}-Bench and the nearest neighbor methods we evaluate. Each dataset is chosen to stress-test matrix completion beyond standard synthetic or recommender-system benchmarks: the tasks span mobile health, recommender systems, causal panel data, and LLM evaluation, and feature structured, adaptive, or confounded missingness patterns. We also support both scalar and distributional matrix completion.

\subsection{Datasets}\label{sec:datasets}

\lib{}-Bench consists of four real-world datasets summarized in \cref{tab:datasets}. Each dataset comes with a data loader that automatically downloads and caches the data, constructs the observation matrix and missingness mask, and provides train-validation-test splits.

\begin{table}[t]
  \caption{\label{tab:datasets}\textbf{Summary of \lib{}-Bench datasets.} $N$ and $T$ denote the matrix dimensions (rows and columns), $n$ denotes the number of measurements per observed entry, and ``Type'' indicates whether the task is scalar ($n=1$) or distributional ($n > 1$) matrix completion.}
  \centering
  \begin{tabular}{llccccl}
    \toprule
    \textbf{Dataset} & \textbf{Domain} & $N$ & $T$ & $n$ & \textbf{Missing \%} & \textbf{Type} \\
    \midrule
    HeartSteps V1 & Mobile health & 37 & 200 & 60 & $\sim$40\% & Scalar + Dist.\ \\
    MovieLens 1M & Recommender systems & 6{,}040 & 3{,}952 & 1 & 95.53\% & Scalar \\
    Proposition 99 & Causal panel data & 39 & 31 & 1 & $\sim$36\% & Scalar \\
    PromptEval & LLM evaluation & 15 & 57 & 100 & varies & Distributional \\
    \bottomrule
  \end{tabular}
\end{table}

\paragraph{HeartSteps V1.} A micro-randomized clinical trial measuring the effect of mobile notifications on physical activity~\cite{klasnja2019efficacy}. $N = 37$ participants were observed over $T = 200$ decision points; each entry records the participant's step count in the hour following a notification. Notifications were sent with probability $p = 0.6$ conditional on availability, inducing a missingness pattern that depends on individual schedules---a violation of the missing-completely-at-random (MCAR) assumption. In the distributional setting, each observed entry contains $n = 60$ one-minute step count samples.

\paragraph{MovieLens 1M.} A standard collaborative filtering benchmark with 1 million ratings (1--5 stars) from 6{,}040 users on 3{,}952 movies~\cite{movielens}. With 95.53\% of entries missing, this is a high-sparsity scalar matrix completion task. The missingness is heavily confounded: popular movies and active users are disproportionately observed.

\paragraph{Proposition 99.} Annual per-capita cigarette sales across 39 U.S.\ states from 1970 to 2000, used to estimate the causal effect of California's 1988 tobacco tax~\cite{abadie2010synthetic}. The missingness follows a \emph{staggered adoption} pattern: California's post-1988 counterfactual consumption is unobserved by definition. This task connects matrix completion to synthetic control methods in causal inference.

\paragraph{PromptEval.} Evaluation of 15 open-source LLMs across 57 MMLU tasks using 100 prompting templates~\cite{polo2024efficient}. Each entry is a distribution of binary scores (correct/incorrect) over the prompt templates, making this a distributional matrix completion task ($n = 100$). The goal is to characterize model performance without exhaustive evaluation.

\paragraph{Benchmark protocol and intended use.} N$^2$-Bench is designed to evaluate matrix completion methods under observation patterns that are common in applications but underrepresented in standard synthetic benchmarks. Each task specifies a matrix construction, an observation mask, a train-validation-test protocol, a prediction target, and a metric. The benchmark is not intended to produce a single universal ranking of methods. Instead, it tests how method behavior changes across four axes: scalar versus distributional entries, random versus structured missingness, recommender style sparsity versus panel counterfactuals, and scalar versus distributional evaluation metrics. Hyperparameters are selected only using validation entries, and final performance is reported on held-out entries or task-specific counterfactual targets.

\subsection{Evaluated methods}\label{sec:nn-gist}

The methods evaluated in \lib{}-\textbf{Bench} include both classical matrix completion methods as well as modern nearest neighbor approaches. For the classical methods, we benchmark two popular baselines: \textbf{SoftImpute}~\cite{hastie2015matrix}, which solves a nuclear-norm regularized optimization, and \textbf{USVT}~\cite{chatterjee2015matrix}, a spectral method based on singular value thresholding. For the Proposition~99 task, we additionally evaluate the synthetic control method of \cite{abadie2010synthetic}.

For the nearest neighbor approaches, we briefly introduce the mathematical model and framework; full details are deferred to \cref{sec:nn-overview}.

\paragraph{NN Model.} We observe a partially observed matrix:
\begin{align}
\qtext{for} i \in [N], t\in[T]:\quad \measurement{i, t} :=
     \begin{cases}
        \sample{1}(i, t), ..., \sample{n}(i, t) \ \sim \ \dist{i, t}&\qtext{if} \missing{i, t}=1, \\
        \mrm{unknown}&\qtext{if} \missing{i, t}=0.
    \end{cases}
    \label{eq:model}
\end{align}
When $n = 1$, this is the \textit{scalar matrix completion} problem, where the goal is to estimate the entry means $\meanparam{i,t} = \int x\, d\dist{i,t}(x)$. When $n \geq 2$, it is the \textit{distributional matrix completion} problem, where the goal is to recover the full distributions $\dist{i,t}$.

\paragraph{NN Framework.} All nearest neighbor methods in \lib{} follow a two-step procedure: (i)~a $\distancemod$ module computes row-wise and column-wise distances between matrix entries using observed data, and (ii)~an $\averagemod$ module computes a weighted average of nearby entries to impute the missing value. Different choices of the distance metric, averaging rule, and weighting scheme yield the variants summarized in \cref{tab:summary}. Of particular note, $\autonn$ is a new variant we introduce that interpolates between $\tsnn$ and $\drnn$ to automatically adapt to the underlying noise level (see \cref{sec:new-nn}).

\setlength\cellspacetoplimit{5pt}
\setlength\cellspacebottomlimit{5pt}

\begin{table}[t]
  \caption{\label{tab:summary}\textbf{Variants of nearest neighbors for matrix completion.}}
  \centering
  \begin{tabular}{clccc}
    \toprule
    \tbf{Type}
    &\tbf{Method}
    & ${\what\varphi}(x, y)$
    & \def\arraystretch{1.2}\begin{tabular}{c} $\varphi(x, y)$
    \end{tabular}
    & $\weight{j, s}$
    \\
    \midrule
   $n = 1$ &
    $\rownn$ \ \citep{li2019nearest} (\cref{algo:vanilla-nn}) & $(x - y)^2$
    & $(x - y)^2$
    & $\indic{\distance{i, j}^\row\leq \tuning_1, \distance{s, t}^\col \leq 0}$
    \\
    &
    $\colnn$ \ \citep{li2019nearest} (\cref{algo:vanilla-nn}) & $(x - y)^2$
    & $(x - y)^2$
    & $\indic{\distance{i, j}^\row\leq 0, \distance{s, t}^\col \leq \tuning_2}$
    \\
    &
    $\tsnn$ \ \citep{sadhukhan2024adaptivity} \textup{(\cref{algo:tsnn})}
     & $(x - y)^2$
     & $(x - y)^2$
    & $\indic{\distance{i, j}^\row\leq \tuning_1, \distance{s, t}^\col \leq \tuning_2}$
    \\
    &
    $\awnn$ \ \citep{sadhukhan2025adaptivelyweightednearestneighborsmatrix} \textup{(\cref{algo:awnn})}
    & $(x - y)^2$
    & $(x - y)^2$
    & $\weight{j}^\star(i, t)\cdot\indic{\distance{s, t}^\col \leq 0}$
    \\
    \noalign{\vspace{5pt}}
    \cline{2-5}
    \noalign{\vspace{5pt}}
    &
    $\drnn$ \ \citep{dwivedi2022doubly} {(\cref{algo:dsnn})}
     & \multicolumn{3}{c}{$\rownn + \colnn - \tsnn$}
    \\
    &
    $\autonn$ \ (\cref{sec:new-nn})
     & \multicolumn{3}{c}{$\alpha \cdot \drnn + (1 - \alpha) \cdot \tsnn$}
    \\
     \midrule
$n > 1$     &
    $\kernelnn$ \ \citep{choi2024learning} (\cref{algo:vanilla-dnn}) & $\what \mmd_\kernel^2(x, y)$
    & $\mmd_\kernel^2(x, y)$
    & $\indic{\distance{i, j}^\row\leq \tuning_1, \distance{s, t}^\col \leq 0}$
    \\
    &
    $\wassersteinnn$ \ \citep{feitelberg2024distributional} (\cref{algo:vanilla-dnn}) & $\hatwasserstein^2(x, y)$
    & $\wasserstein^2(x, y)$
    & $\indic{\distance{i, j}^\row\leq \tuning_1, \distance{s, t}^\col \leq 0}$
    \\ \bottomrule
  \end{tabular}
\end{table}

%% file: neurips2026_tmp/case-study.tex
\section{\lib{}-Bench and Results}\label{sec:case-study}


We now evaluate matrix completion methods on the four \lib{}-\textbf{Bench} tasks. The goal is not to produce a single universal leaderboard, but to test how methods behave across realistic missingness patterns, domains, and entry types. The benchmark includes scalar prediction tasks, distributional prediction tasks, and causal panel-data tasks. For each task, \lib{}-\textbf{Bench} specifies the matrix construction, observation mask, validation protocol, metric, and set of baselines. We compare nearest-neighbor methods against spectral, optimization-based, and causal inference baselines when applicable. Additional preprocessing and implementation details are provided in \cref{app:case-studies}.

\subsection{Personalized healthcare: HeartSteps}
\label{sub:hs}


The HeartSteps V1 study~(HeartSteps study for short) is a clinical trial designed to measure the efficacy of the HeartSteps mobile application for encouraging non-sedentary activity \cite{klasnja2019efficacy}. The HeartSteps V1 data and its subsequent extensions have been widely used for benchmarking a variety of tasks including counterfactual inference of treatment effect \cite{dwivedi2022counterfactual, choi2024learning}, reinforcement learning for intervention selection \cite{liao2019rlheartsteps}, and micro-randomized trial design \cite{qian2022microrandomized}. In the HeartSteps study, $N = 37$ participants were under a 6-week period micro-randomized trial, where they were provided with a mobile application and an activity tracker. Participants independently received a notification with probability $p = 0.6$ for $5$ pre-determined decision points per day for 40 days~($T = 200$). We denote observed entries $Z_{i, t}$ as the mean participant step count for one hour after a notification was sent and unobserved entries as the unknown step count for decision points where no notification was sent. Our task is to estimate the counterfactual outcomes: the participant's step count should they have received a different treatment (notification or no notification) than they did at specific time points during the study. 





\paragraph{Results \& Discussion.} We benchmark the performance of the matrix completion methods by measuring absolute error on held-out observed step counts across 10 participants in the last 50 decision points. We use the remaining data to find nearest neighbor hyperparameters using cross-validation. To benchmark the distributional nearest neighbors methods ($\kernelnn$ and $\wassersteinnn$) against the scalar methods, we first set each entry to have the number of samples $n = 60$, where each sample is the 1 minute step count before imputation. Then, we take the mean of the imputed empirical distribution as the estimate. 

\begin{figure}[t]
    \centering
    \begin{tabular}{cc}
        \includegraphics[width=0.47\textwidth]{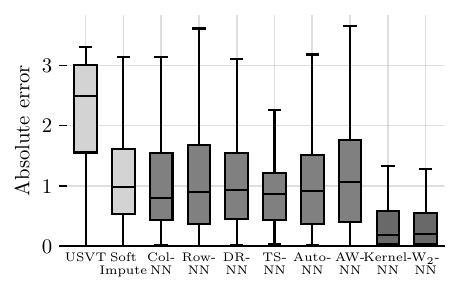} &  
        \includegraphics[width=0.33\textwidth]{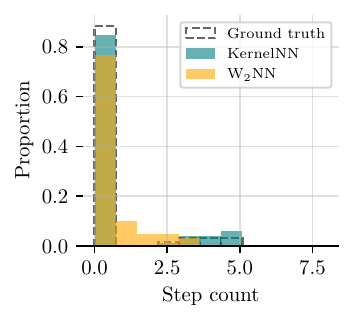} \\
        (a) Absolute error of mean step count prediction &
        (b) $\kernelnn$ vs. $\wassersteinnn$
    \end{tabular}
    \caption{\tbf{HeartSteps: estimating step count under scalar and distributional matrix completion settings.} Panel (a) shows the absolute error of predicted step count of the nearest neighbor methods against matrix completion baselines (SoftImpute, USVT). Panel (b) shows an example of an imputed entry in the distributional matrix completion setting.
    }
    \label{fig:hs-results}
\end{figure}

In \cref{fig:hs-results}(a), we compare the absolute error of the imputed values across the nearest neighbor and baseline methods. The scalar nearest neighbor methods far out-perform USVT and are on par with SoftImpute. The two distributional nearest neighbor methods far outperform all methods operating in the scalar setting; it suggests that matching by distributions collect more homogeneous neighbors, thereby decreases the bias of the method, compared to matching only the first moments as done in most scalar matrix nearest neighbor methods. 


In \cref{fig:hs-results} panel (b), we show an example of an imputed entry in the distributional nearest neighbors setting. In this case, the ground truth distribution is bimodal, as the participant was largely sedentary (0 steps) with small amounts of activity. While both $\kernelnn$ and $\wassersteinnn$ capture the sedentary behavior of the participant, $\kernelnn$ is able to recover the bimodality of the original distribution whereas $\wassersteinnn$ cannot.


\subsection{Movie recommendations: MovieLens}\label{sec:movielens}
\begin{wrapfigure}{R}{0.5\textwidth}
    \begin{center}        \includegraphics{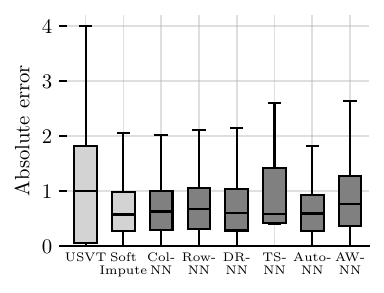}
    \end{center}
    \caption{\tbf{MovieLens: Estimation error for a random subsample of size 500.} For experimental settings and discussion see \cref{sec:movielens}.}
    \label{fig:movielens-error}
\end{wrapfigure}

The MovieLens 1M dataset \cite{movielens} contains 1 million ratings (1–5 stars) from 6,040 users on 3,952 movies. Collaborative filtering on MovieLens has long been a benchmark for matrix-completion methods: neighborhood-based algorithms~\cite{sarwar2001item}, latent-factor models~\cite{koren2009matrix}, and, more recently, nearest neighbors interpreted as blind regression under a latent–variable model~\cite{li2019nearest}. These assist practitioners in data-driven recommendation systems, since more accurate rating imputation directly drives better personalized suggestions and user engagement. This is a standard scalar matrix completion problem with $N=6,040$ and $T=3,952$. Each rating is an integer in $\{1,\dots,5\}$. The dataset has a very high percentage of missing values: 95.53\% missing. Our task is to estimate unobserved ratings using various matrix completion algorithms. We benchmark the performance of nearest neighbors against matrix factorization by measuring absolute error on held‑out ratings. See \cref{app:movielens} for additional details on the dataset.

\paragraph{Results \& Discussion.} We fit the nearest neighbor methods using a random sample of size 100 from the first 80\% of the dataset to choose nearest neighbor hyperparameters via cross-validation.We then test the method on a random subsample of size 500 from the last 20\% of the dataset. As observed in \cref{fig:movielens-error}, all nearest neighbor methods have a lower average error than USVT and a much lower standard deviation of errors, with $\colnn$, $\rownn$, $\drnn$, and $\autonn$ performing the best out of the nearest neighbor methods. SoftImpute performs on par with the nearest neighbor methods. Note that the nearest neighbor methods perform well even while only being trained on a tiny subset of the data of size 100 out of the 1 million ratings available.

\subsection{Counterfactual inference for panel data: Proposition 99}
\label{sub:prop99}

\begin{figure}[t]
    \centering
    \begin{tabular}{cc}
        \includegraphics[width=0.42\textwidth]{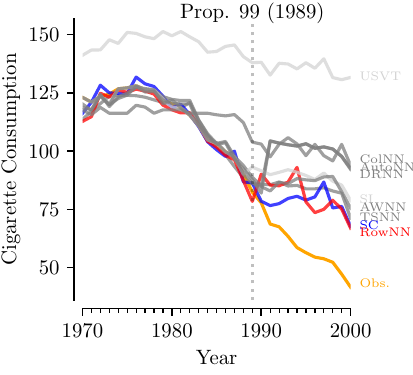} &
        \includegraphics[width=0.47\textwidth]{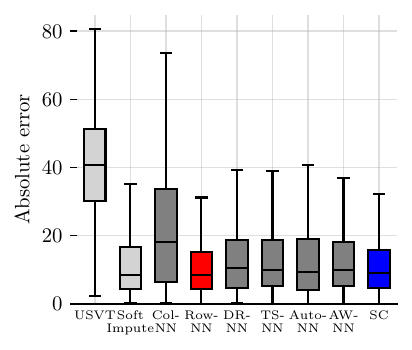} \\
        (a) Synthetic controls for California &
        (b) Absolute error on control states \\ in post-intervention period
    \end{tabular}
    \caption{\tbf{Nearest neighbor methods generate high-fidelity synthetic controls in counterfactual inference for panel data.} For exact settings and further discussion see \cref{sub:prop99}.
    }
    \label{fig:prop99}
\end{figure}

Next we consider a panel data setting, where our goal is to estimate the effect of the California Tobacco Tax and Health Protection Act of 1988 (a.k.a. Proposition 99) on annual state-level cigarette consumption\footnote{measured as per capita cigarette sales in packs}.
By definition, the counterfactual cigarette consumption in California---had Proposition 99 never been enacted---is not observed. 
\cite{abadie2010synthetic} introduce the notion of a ``synthetic control'' to serve as a proxy for this unobserved value based on ``neighboring'' \emph{control states} that never instituted a tobacco tax. These states are not close in a geographical sense, but rather close due to similarities in other covariates\footnote{GDP per capita, beer consumption, percent aged 15–24, and cigarette retail prices}. 
We take a different approach and use only the observed cigarette consumption levels from the control states, of which there are 38 in total. Thus, we frame our problem as a scalar matrix completion problem with $N=39$ and $T=31$ (see Eq. \cref{eq:model}). The last row in the matrix corresponds to the state of California. 

\paragraph{Results \& Discussion.}
For each method, we use a 64-16-20 train-validation-test split and use cross validation to fit any hyperparameters. \cref{fig:prop99} plots the various synthetic controls for California (left) and absolute error of each method on the 38 control states, for which we do observe the no-treatment values (right).
From \cref{fig:prop99}(a), we see that nearest neighbor methods, in particular $\tsnn$ and $\rownn$, are roughly on par with the gold-standard synthetic control method of \cite{abadie2010synthetic} (``SC'') for estimating California's counterfactual cigarette consumption in the post-intervention period (after 1989). This is despite the fact that the nearest neighbor methods rely on less information for the estimation task.
From \cref{fig:prop99}(b), we see that all nearest neighbor methods, with the exception of $\colnn$, achieve similar error levels as the synthetic control baseline. $\rownn$ achieves even lower error levels.
See supplementary experiment details in \cref{app:prop99}.







%% file: neurips2026_tmp/llm_evaluation_new.tex
\subsection{Efficient LLM evaluation: PromptEval}
\label{sub:llm}


The rapid advancement of LLMs have placed them at the center of many modern machine learning systems, from chatbots to aids in medical education \cite{geathers2025benchmarking}. In practice, system architects want to strike the right balance of real-world performance and cost, but navigating this Pareto frontier is a daunting task. 2024 alone saw at least 10 new models from Anthropic, Google, Meta, and OpenAI, not even counting the multitude of open-source fine-tuned models built on top of these.
On specific tasks, smaller, fine-tuned models may even outperform the latest frontier models, in addition to being more cost effective.

We investigate how matrix completion, specifically nearest neighbor methods, can alleviate some of these burdens. We use the PromptEval dataset \cite{polo2024efficient}, which evaluates $15$ open-source language models (ranging in size from 3B to 70B parameters) and $100$ different prompting techniques across the 57 tasks of the MMLU benchmark \citep{hendrycks2020measuring}. In practice, the performance of a model depends---sometimes dramatically---on the precise input prompt. This suggests that we need to consider the performance of a model across a wide range of prompts, rather than any one prompt in particular. 
Thus, we model this problem as a distributional matrix completion problem with $N=15$, $T=57$, and $n=100$.
Given one of 57 tasks, we aim to accurately characterize the performance of each model without resorting to exhaustive evaluation.
Nearest neighbors leverage commonalities across models and tasks to estimate the performance distribution of each entry, which was otherwise not considered in \cite{polo2024efficient}; previous literature achieves efficient evaluation per model and task in isolation without leveraging any across model / task information. 

\begin{figure}[t]
    \centering
    \begin{tabular}{cc}
        \includegraphics[width=0.47\textwidth]{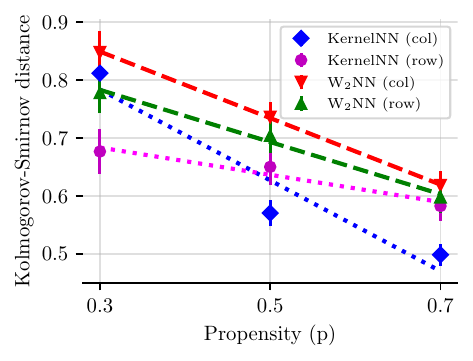} & \includegraphics[width=0.45\textwidth]{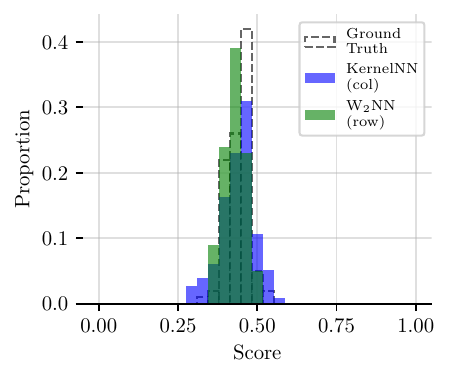} \\
        (a) Mean KS distance between estimated &
        (b) $\kernelnn$ vs. $\wassersteinnn$ \\
        and ground-truth distributions
    \end{tabular}
    \caption{\textbf{Distributional nearest neighbor methods enable efficient LLM evaluation on MMLU.} We estimate LLM score distributions across all models and tasks given only a limited number of model-task evaluations, determined by the propensity $p$. See \cref{sub:llm} for a detailed discussion.}
    \label{fig:llm}
\end{figure}

\paragraph{Results \& Discussion.}
We randomly include each entry in the matrix independently with probability $p\in \braces{0.3,0.5,0.7}$ and impute the missing entries using the $\kernelnn$ and $\wassersteinnn$ methods of \cref{tab:summary}. For each method, we consider both the the row-wise and column-wise variants.
\cref{fig:llm}(a) reports the mean Kolmogorov-Smirnov (KS) distance between the imputed and ground-truth distributions across the entries in the test set for varying missingness values. As expected, estimation error decreases as $p$ increases.
\cref{fig:llm}(b) visualizes the imputed distributions using row-wise $\kernelnn$ and column-wise $\wassersteinnn$ (at $p=0.7$) for a select entry, along with the ground-truth distribution. Even with 30\% of matrix entries missing, distributional NN methods are able to recover the underlying distribution.

\paragraph{What \lib{}-\textbf{Bench} reveals.} Across the four tasks, method performance depends strongly on the observation pattern and entry type. On HeartSteps, distributional nearest-neighbor methods improve scalar prediction when multiple samples are available per entry, suggesting that the full empirical distribution can be a more informative matching signal than the mean alone. On MovieLens, nearest-neighbor methods are competitive with SoftImpute and substantially stronger than USVT on the sampled held-out ratings. On Proposition 99, row-wise and two-sided NN methods recover counterfactual trajectories close to synthetic control while using only the outcome panel rather than additional covariates. On PromptEval, distributional NN methods recover model-task score distributions from partial evaluations, with KS distance decreasing as the observation probability increases. Together, these results suggest that matrix completion benchmarks should include structured missingness, causal panels, and distributional entries rather than relying only on synthetic low-rank matrices or recommender-system data.

%% file: neurips2026_tmp/library.tex
\section{\lib{} Package and Interface}\label{sec:library}

We now present our unified Python package, \lib{}, for nearest neighbor algorithms for matrix completion. In particular, we provide a class structure which abstracts the estimation procedure utilized in each different nearest neighbor method as well as the $\distancemod$ and $\averagemod$ modules described in \cref{sec:nn-gist} (see \cref{sec:nn-overview} for full details). On top of that, our library facilitates easy extensions to other nearest neighbors algorithms and other data types on top of scalars and distributions. For example, as long as a distance and average notion are well defined, our library can be easily applied to a matrix of images or text strings.

\paragraph{Class structure.} The core functionality of \lib{} is based on two abstract classes: \texttt{EstimationMethod} and \texttt{DataType}. 

\texttt{EstimationMethod} classes contain the logic to impute a missing entry such as how to use calculated distances. In the context of the modules in \cref{sec:nn-gist}, this class determines the weighting function or, in the context of $\drnn$ and $\autonn$, how to compose estimates. We separate this from the \texttt{DataType} abstraction because several estimation methods can be used for multiple data types. For example, \texttt{RowRowEstimator} implements the $\rownn$ procedure for any data type given to it, such as scalars or distributions.

\texttt{DataType} classes implement the $\distancemod$ and $\averagemod$ modules for any kind of data type (e.g. scalars which use squared distance and simple averaging). This abstract class allows for our package to extend to any data types beyond the ones we tested. For instance, a practitioner can easily add a \texttt{DataType} for text strings which uses vector embeddings to find distances and averages between between strings without needing to rewrite any of the estimation procedure.

\paragraph{Interface.} To use our library, a user simply has to instantiate a composite class \texttt{NearestNeighborImputer} with their \texttt{EstimationMethod} and \texttt{DataType} of choice. We provide constructor functions to automatically create popular \texttt{NearestNeighborImputer} classes such as a two-sided nearest neighbor estimator with the scalar data type. From a design pattern point of view, this is known as a \emph{Composite} design pattern \cite[pg. 163]{gamma1993design}. We use this design pattern so that anyone looking to customize the estimation procedure can do so for any kind of data type simultaneously. Similarly, with the exception of doubly robust estimators, each estimation procedure works out of the box with any data type that implements the \texttt{DataType} abstract class. The Doubly robust estimation method does not work out of the box with distributions because a subtraction operation is not well defined in the distribution space.

Finally, the user simply needs to input (i) a data matrix, (ii) a mask matrix which specifies which values are missing, and (iii) the row and column to impute. Thus, a user can test out different estimation procedures by changing just one line of code. Separately from the core functionality, we have also implemented several cross-validation classes which take in a \texttt{NearestNeighborImputer} class and find the best hyperparameters to use (e.g., distance thresholds and weights).





%% file: neurips2026_tmp/conclusion.tex
\section{Conclusion}\label{sec:conclusion}

In this paper, we introduce \lib{}\textbf{-Bench}, a benchmark suite for matrix completion on realistic data spanning mobile health (HeartSteps), recommender systems (MovieLens), causal panel data (Proposition 99), and LLM evaluation (PromptEval), supporting both scalar and distributional settings. Alongside the benchmark, we release \lib{}, a modular Python library that unifies a broad family of nearest neighbor methods—row-wise, column-wise, two-sided, doubly robust, adaptively weighted, and distributional variants—under a common interface, and we introduce AutoNN, a new variant that automatically interpolates between debiasing and denoising regimes.

Our empirical results demonstrate that conclusions drawn from idealized synthetic experiments can be misleading on real data. Spectral and optimization-based methods such as USVT and SoftImpute, which enjoy strong guarantees under random missingness, often degrade when missingness is structured, adaptive, or confounded. In contrast, nearest neighbor methods are consistently competitive and notably more stable across domains, matching synthetic control on Proposition 99, outperforming USVT on MovieLens with as few as 100 training ratings, and, in their distributional variants, recovering bimodal step-count distributions on HeartSteps and LLM performance distributions on PromptEval that scalar methods cannot capture.

Several future directions are natural. The benchmark can be extended to additional domains and missingness regimes, and richer weighting schemes beyond those studied here can be incorporated into the library. Scaling to larger matrices will require handling distributed datasets and accelerating N² via efficient matrix-multiplication-based implementations or approximate nearest neighbor strategies. We hope N²-Bench provides a foundation for more realistic and reproducible evaluation of matrix completion methods going forward.


%% file: neurips2026_tmp/nn-overview.tex
\section{Nearest Neighbors for Matrix Completion: Full Details}\label{sec:nn-overview}

This appendix provides the full technical details of the nearest neighbor framework summarized in \cref{sec:nn-gist}. Recall the matrix completion model \cref{eq:model}: for matrix entries where $\missing{i, t} = 1$, we observe $n$ measurements $\measurement{i, t}$ that takes value $\sample{i, t}$ realized from distribution $\dist{i, t}$. When $n = 1$, i.e., $\measurement{i,t} = \sample{1}(i, t)$, we refer to \cref{eq:model} as the \textit{scalar matrix completion} model; scalar matrix completion is the most common problem posed in the literature~\cite{candes2012exact,recht2011simpler,koren2009matrix,hastie2015matrix,chatterjee2015matrix,dwivedi2022counterfactual,dwivedi2022doubly,agarwal2023causal}, where the goal is to learn the mean of the underlying distributions $\sbraces{ \meanparam{i, t} = \int x d\dist{i, t}(x)}_{i \in [N],t \in [T]}.$ When there are more than one observed measurements per entry, i.e., $\measurement{i, t} = [\sample{1}(i, t), ..., \sample{n}(i, t)]$ for $n \geq 2$, we refer to \cref{eq:model} as the \textit{distributional matrix completion} problem, the goal being the recovery of the distributions as a whole.
We refer the readers to \cref{app:guarantees} for a detailed discussion on the structural assumptions imposed on the model \cref{eq:model}.



\subsection{Unified framework}\label{sec:unified-view}
We introduce two general modules~(namely $\distancemod$ and $\averagemod$) from which the variants of nearest neighbors are constructed. We introduce several shorthands used in the modules. Denote the collection of measurements, missingness, and weights:
\begin{align}
    \mc Z:=\big\{\measurement{j, s}\big\}_{j \in [N] , s \in [T]} , \quad  \mc A := \sbraces{\missing{j, s}}_{j \in [N], s \in [T]}, \qtext{and} \mc W := \sbraces{w_{j, s}}_{j \in [N],s \in [T]}.
\end{align} 
Let $\entrydist{x, x'}$ be a metric between $x, x' \in \mc X$ for some space $\mc X$. Further define $\entrydistest{\measurement{i, t}, \measurement{j, s}}$ as a data-dependent distance between any two observed entries $(i, t)$ and $(j, s)$ of the matrix \cref{eq:model}. The two modules can now be defined: 
\begin{itemize}
    \item[(i)] $\distancemod (\what{\varphi}, \mc Z, \mc A)$: Additional input is the data-dependent distance between entries of matrix $\what{\varphi}$ and output is the collection of row-wise and column-wise distance of matrix:
    \begin{align}\label{mod:distance}
        \distance{i, j}^{\row} := \frac{\sum_{s \neq t}\missing{i, s}\missing{j, s}\entrydistest{\measurement{i, s}, \measurement{j, s}}}{\sum_{s \neq t} \missing{i, s}\missing{j, s}} \qtext{and} \distance{t, s}^{\col} := \frac{\sum_{j \neq i}\missing{j, t}\missing{j, s}\entrydistest{\measurement{j, t}, \measurement{j, s}}}{\sum_{j \neq i} \missing{j, t}\missing{j, s}},
    \end{align}
    \item[(ii)] $\averagemod (\varphi, \mc W, \mc Z, \mc A)$: Additional input are the weights $\mc W$, metric $\varphi$ and output is the optimizer
        \begin{align}\label{eq:ave-output}
           \what{\theta} = \argmin_{x \in \mc X}\sum_{j \in [N], s \in [T]} \weight{j, s}\missing{j, s}\entrydist{x, \measurement{j, s}}.
        \end{align}
\end{itemize}
The $\distancemod$ module calculates the row-wise and column-wise distance of the matrix, by taking the average of the observed entry-wise distance $\entrydistest{\cdot, \cdot}$. The $\averagemod$ module calculates the weighted average of observed measurements, where the notion of average depends on the metric $\varphi$ and the space $\mc X$ on which the metric $\varphi$ is defined. 
Notably, the weights $\mc W$ in the $\averagemod$ module encodes the entry information of the estimand. 

\begin{remark}\label{rem:row-nn}
The vanilla row-wise nearest neighbors~\cite{li2019nearest} that targets the mean $\theta_{i, t} = \int x d\dist{i, t}(x)$ of entry $(i, t)$ is recovered by first applying $\distancemod$ with $\entrydistest{\measurement{j, s}, \measurement{j', s'}} = {(\measurement{j, s} - \measurement{j', s'})^2}$, applying $\averagemod$ with the non-smooth weight
${\weight{j, s} = \indic{\distance{i,j}^\row \leq \tuning_1}\cdot \indic{\distance{s, t}^\col \leq 0}}$, and using the metric $\entrydist{x, y} = (x - y)^2$. Note that the non-smooth weight satisfies $\weight{j, t} = \indic{\distance{i, j}^\row \leq \tuning_1}$, whereas $\weight{j, s} = 0$ for $s \neq t$; by defining the nearest neighbor set $\mbf N_{t, \tuning_1} := \{ j \in [N] : \distance{j, t}^\row \leq \tuning_1 \}$, the $\averagemod$ module output can be rewritten as ${\argmin_{x \in \real} \sum_{j \in \mbf N_{t, \tuning_1}} \missing{j, t}(x - \measurement{j, t})^2 = |\mbf N_{t, \tuning_1}|^{-1}\sum_{j \in \mbf N_{t, \tuning_1}}\measurement{j, t}}$.
\end{remark}

\subsection{Existing methods}
\label{subsec: existing methods}
We present existing variants of nearest neighbors using the two modules introduced in \cref{sec:unified-view}; all the methods summarized in \cref{tab:summary} are recovered by sequentially applying $\distancemod$ and $\averagemod$ with the appropriate specification of $\what \varphi$, $\varphi$ and $\mc W$. 

All methods except $\awnn$ and our newly proposed $\autonn$, have binary weights i.e., $w_{j,s}\in\sbraces{0,1}$. $\autonn$, detailed in \cref{sec:new-nn}, uses weights to carefully pool together the benefits of $\tsnn$ and $\drnn$. $\awnn$~\cite{sadhukhan2025adaptivelyweightednearestneighborsmatrix} improves upon $\rownn$  by adaptively choosing the weights which optimally balances the bias-variance tradeoff of $\rownn$ as follows 
\begin{align}\label{eq:star-nn-weight}
  \big(\weight{1}^\star(i, t), ..., \weight{N}^\star(i, t)\big):= \argmin_{(v_{1}, ..., v_{N})\in \Delta_N} 2\log(2N)\widehat \sigma^2 \sum_{k \in [N]}v_{k}^2 + \sum_{k \in [N]}v_{k} A_{k,t}\distance{i,k}^\row  .
\end{align}
where $\widehat \sigma^2$ is the estimated error and $\Delta_N$ is a simplex in $\real^N$; see \cite{sadhukhan2025adaptivelyweightednearestneighborsmatrix} for details of \cref{eq:star-nn-weight}. \cref{tab:summary} contains a concise summary of the existing nearest neighbor variants; see \cref{app:nn-algo} for a detailed exposition for each methods.

Under the distributional matrix completion setting~($n > 1$ in \cref{eq:model}), the methods $\kernelnn$ and $\wassersteinnn$ in \cref{tab:summary} take $\mu, \nu \in \mc X$ as square integrable probability measures, and $\entrydist{\mu, \nu}$ as either the squared maximum mean discrepency~(i.e. $\mmd_\kernel^2(\mu, \nu)$, see \cite{muandet2017kernel}) or squared Wasserstein metric~(i.e., $\wasserstein(\mu, \nu)$, see \cite{bigot2020}).
Further, the entry-wise distance $\entrydistest{x, y}$ in this case is either the unbiased U-statistics estimator $\what{\mmd}_\kernel^2(\measurement{i, t}, \measurement{j, s})$ for $\mmd_\kernel^2(\dist{i, t}, \dist{j, s})$~(see \cite{muandet2017kernel})
or the quantile based estimator $\hatwasserstein(\measurement{i, t}, \measurement{j, s})$ for $\wasserstein(\dist{i, t}, \dist{j, s})$~(see \cite{bigot2020}).

%% file: neurips2026_tmp/nn-new.tex
\subsection{New variant: Auto nearest neighbors}\label{sec:new-nn}

$\tsnn$ is a generalization of $\rownn$ and $\colnn$ by setting one of the tuning parameters to zero~(see \cref{tab:summary}), whereas the idea underlying $\drnn$ is fundamentally different from that of $\tsnn$; $\drnn$ debiases a naive combination of $\rownn$ and $\colnn$ whereas $\tsnn$ simply boosts the number of measurements averaged upon, thereby gaining from lower variance. 
So we simply interpolate the two methods 
for some hyper-parameter $\alpha \in [0, 1]$; see \cref{tab:summary}. 
Notably the hyper-parameter $\tuning$ for both $\drnn$ and $\tsnn$ are identical when interpolated.

\begin{figure}[H]
    \centering
    \begin{tabular}{cc}
        \includegraphics[width=0.43\textwidth]{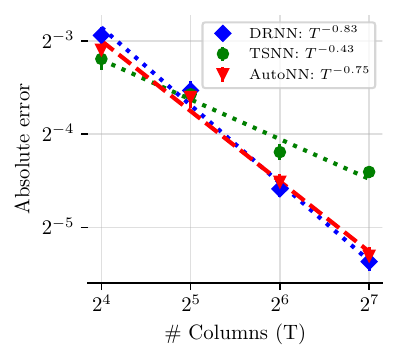} &  
        \includegraphics[width=0.43\textwidth]{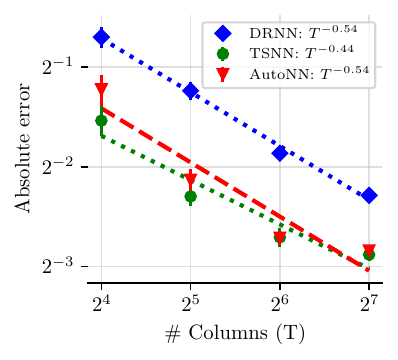}
        \\ (a) High SNR & (b) Low SNR
    \end{tabular}
    \caption{\tbf{Error scaling for certain NN variants in synthetic experiments.}
    See \cref{app:synthetic_gen} for details on the data-generating process and how the signal-to-noise ratio (SNR) is defined. Each point corresponds to the mean absolute error ± 1 standard error across 30 trials.}
    \label{fig:auto_perf}
\end{figure}

Suppose $\dist{i, t} = \theta_{i, t} + \varepsilon_{i, t}$ in \cref{eq:model} where $\varepsilon_{i, t}$ are centered i.i.d. sub-Gaussian distributions across $i $ and $t$.
When $\sigma$ is large in magnitude, $\tsnn$ denoises the estimate by averaging over more samples, hence providing a superior performance compared to $\drnn$ in a noisy scenario. When $\sigma$ is small so that bias of nearest neighbor is more prominent, $\drnn$ effectively debiases the estimate so as to provide a superior performance compared to $\tsnn$. The linear interpolator $\autonn$ \textit{automatically} adjusts to the underlying noise level and debiases or denoises accordingly; such property is critical when applying nearest neighbors to real world data set where the noise level is unknown. We refer to \cref{fig:auto_perf} for visual evidence.



\newcommand{\colsize}{N_{1, \eta_2}^{\mrm{col}}}
\newcommand{\rowsize}{N_{1, \eta_1}^{\mrm{row}}}

%% file: appendices/guarantees.tex
\section{Structural assumptions}
\label{app:guarantees}

Provable guarantees of nearest neighbors in matrix settings \cref{eq:model} can be shown when structural assumptions are imposed on the distributions $\dist{i, t}$ and the missingness $\missing{i, t}$. We collect existing results from \cite{li2019nearest,dwivedi2022doubly,choi2024learning,feitelberg2024distributional,sadhukhan2024adaptivity,sadhukhan2025adaptivelyweightednearestneighborsmatrix}. Given data with missing observations from \cref{eq:model}, the practitioner is interested in learning information of the distributions, e.g., mean of the distributions $\sbraces{ \meanparam{i, t} = \int x d\dist{i, t}(x)}.$



The first assumption specifies the factor structure on the mean; that is, there exists latent factors $u_i, v_t$ that collectively characterize the signal of each entry $(i, t)$ of the matrix~\cite{li2019nearest,dwivedi2022counterfactual,agarwal2023causal,choi2024learning,feitelberg2024distributional}. Such a factor model is analogous to the low rank assumptions commonly imposed in matrix completion~\cite{candes2012exact}. The second assumption specifies how the missing pattern $\missing{i, t}$ was generated; for instance missing completely at random~(MCAR) assumes that $\missing{i, t}$ are independent to all other randomness present in the model \cref{eq:model} and that all entries have positive probability of being observed.


\subsection{Factor model}

For the scalar matrix completion problem, i.e., \cref{eq:model} with $n = 1$, the main goal is to learn~(or impute) the mean of the underlying distribution $\theta_{i, t}$ for any missing entries~\cite{li2019nearest,dwivedi2022doubly,dwivedi2022counterfactual,agarwal2023causal,sadhukhan2024adaptivity, sadhukhan2025adaptivelyweightednearestneighborsmatrix}. The majority of this literature assumes (i) an additive noise model $\dist{i, t} = \theta_{i, t} + \varepsilon_{i, t}$ for centered i.i.d. sub-Gaussian noise $\varepsilon$ and (ii) mean factor model, i.e., $\theta_{i, t} = f(u_i, v_t)$ for some latent factors $u_i, v_t$ and real valued function $f$. 

For the distributional matrix completion problem (i.e., \cref{eq:model} with $n > 1$) the main goal is to learn the underlying distribution itself~\cite{choi2024learning,feitelberg2024distributional}; a factor model is imposed on the distribution as a whole. For instance, a factor model is assumed on the kernel mean embedding of distributions; that is, there exist latent factors $u_i$ and $v_t$ and an operator $g$ such that $\int \kernel(x, \cdot)d\dist{i, t}(x) = g(u_i, v_t)$.

\subsection{Missingness pattern}
\label{app:missingness}

For both the scalar and distributional matrix completion problem \cref{eq:model}, the missing pattern~(i.e., how the missingness $\missing{j, s}$ was generated) can be categorized into three classes using the taxonomy of \cite{little2019statistical}: missing-completely-at-random~(MCAR), missing-at-random~(MAR) and missing-not-at-random~(MNAR). MCAR assumes that the missingness $\missing{i, t}$ is exogenous~(independently generated from all the randomness in the model) and i.i.d. with propensity $\Prob(\missing{i, t} = 1) = p > 0$ for all $(i, t)$. MAR is a more challenging scenario compared to MCAR as missingness is not exogenous, but its randomness depends on the observations. Further, propensities $p_{i, t}$ may differ for entries $(i, t)$ but positivity still holds, i.e., $\min_{i \in [N], t \in [T]}p_{i, t} > 0$. An important instance for MAR is the adaptive randomized policies~\cite{dwivedi2022personalization}. The MNAR setup is the most challenging as it assumes the missingness depends on the unobserved latent confounders, while positivity may also be violated, i.e., $\min_{i \in [N], t \in [T]}p_{i,t } = 0$. The staggered adoption pattern, where a unit remains treated once a unit is treated at some adoption time, is a popular example of MNAR, mainly because positivity is violated. See \cite{athey2021matrix,athey2022design} for more details on staggered adoption.

We briefly outline the structural assumptions existing nearest neighbor methods were shown to work with provable guarantees; for all the existing methods, factor models~(with slightly different details; compare the mean factorization~\cite{li2019nearest} and the distribution factorization~\cite{choi2024learning,feitelberg2024distributional}) are all commonly assumed.
\begin{itemize}
    \item (Scalar matrix completion) The vanilla versions of nearest neighbors~($\rownn$) in \cite{li2019nearest,dwivedi2022counterfactual} are shown to work for MCAR and MAR setup; the latter shows that simple nearest neighbors can provably impute the mean when the missingness is fully adaptive across all users and history. The variants of vanilla nearest neighbors $\drnn$~\cite{dwivedi2022doubly} is proven to work under MCAR, while $\tsnn$~\cite{sadhukhan2024adaptivity} is proven to work under unobserved confounding, i.e., MNAR. 
    \item (Distributional matrix completion) The $\kernelnn$~\cite{choi2024learning} is shown to recover the underlying distribution under MNAR, whereas $\wassersteinnn$~\cite{feitelberg2024distributional} is shown to work under MCAR. 
\end{itemize}








%% file: appendices/nn-algo.tex
\section{Nearest neighbor algorithms}\label{app:nn-algo}

The nearest neighbor methods introduced in \cref{tab:summary} are elaborated in this section. We present two versions of each method; the first version explicitly constructs neighborhoods instead of subtly embedding them in the weights $\mc W$ of the $\averagemod$ module, and the second version specifies how each methods can be recovered by applying the two modules, $\distancemod$ and $\averagemod$, sequentially.




\subsection{Vanilla nearest neighbors}

We elaborate on the discussion in \cref{rem:row-nn} and provide here a detailed algorithm based on the explicit construction of neighborhoods, which is essentially equivalent to $\rownn$ in \cref{tab:summary}. The inputs are measurements $\mc Z$, missingness $\mc A$, the target index $(i, t)$, and the radius $\tuning$.
\begin{itemize}
    \item[\textbf{Step 1:}] (Distance between rows) Calculate the distance between row $i$ and any row $j \in [N] \setminus \{i\}$ by averaging the squared Euclidean distance across overlapping columns:
    \begin{align}\label{eq:snn-distance}
        \distance{i, j} := \frac{\sum_{s \neq t}\missing{i, s}\missing{j, s}(\measurement{i, s} - \measurement{j, s})^2}{\sum_{s \neq t} \missing{i, s}\missing{j, s}}.
    \end{align}
    \item[\textbf{Step 2:}] (Construct neighborhood) Construct a neighborhood of radius $\tuning$ within the $t$th column using the distances $\{\distance{i, j}:j \neq i\}$: 
    \begin{align}
        \neighbors{t, \tuning} := \big\{ j \in [N] \setminus \{i\} : \distance{i, j} \leq \tuning \big\}
    \end{align}
    \item[\textbf{Step 3:}] (Average across observed neighbors) Take the average of measurements within the neighborhood:
    \begin{align}\label{eq:snn-average}
        \what{\theta}_{i, t, \eta} := \frac{1}{|\neighbors{t, \tuning}|}\sum_{j \in \neighbors{t,\tuning}}\missing{j, t}\measurement{j, t}.
    \end{align}
\end{itemize}
In practice, the input $\tuning$ for $\rownn$ should be optimized via cross-validation; we refer the reader to \cref{app:cv} for a detailed implementation.

We specify the exact implementation of the two modules $\distancemod$, $\averagemod$ to recover $\rownn$:
\begin{algorithm2e}[H]
\caption{$\rownn$ for scalar nearest neighbor} 
  \label{algo:vanilla-nn}
  \SetAlgoLined
  \DontPrintSemicolon
  \small
  {
\KwIn{\textup{$\mc Z, \mc A, \tuning,$ $(i, t)$}}
  \BlankLine
    Initialize entry-wise metric $\entrydistest{\measurement{j, s}, \measurement{j', s'}}\gets ( \measurement{j, s} - \measurement{j', s'} )^2$ and metric $\entrydist{x, y} \gets (x - y)^2$\\
    Initialize hyper-parameter ${\tuning} \gets (\tuning_1, 0)$\\
    Calculate row-wise metric $\big\{ \distance{i, j}^\row: j \neq i \big\} \gets $ $\distancemod(\what{\varphi}, \mc Z, \mc A)$ \\
    Initialize weight $\weight{j, s} \gets  \indic{\distance{i, j}^\row \leq \tuning_1, \distance{s, t}^\col \leq \tuning_2}$ \\
    Calculate average $\what \theta_{i, t} \gets \averagemod(\varphi, \mc W, \measurement{}, \missing{})$
    \BlankLine
\KwRet{$\what \theta_{i, t}$}
}
\end{algorithm2e} 
The discussion for $\rownn$ here can be identically made for $\colnn$ as well.

\subsection{Two-sided and doubly-robust nearest neighbors}


We elaborate on the variants of the vanilla nearest neighbors algorithm $\tsnn$ and $\drnn$ in \cref{tab:summary}; we first elaborate on an equivalent version of each of the methods which explicitly constructs neighborhoods.

In the following three step procedure, $\drnn$ and $\tsnn$ differs in the last averaging step: the inputs are the measurements $\mc Z$, missingness $\mc A$, the target index $(i, t)$, and the radii $\tuning = (\tuning_1, \tuning_2)$.
\begin{itemize}
    \item[\textbf{Step 1:}] (Distance between rows) Calculate the distance between row $i$ and any row $j \in [N] \setminus \{i\}$ and the distance between column $t$ and any column $s \in [T] \setminus \{t\}$:
    \begin{align}\label{eq:snn-distance}
        \distance{i, j}^{\row} := \frac{\sum_{s \neq t}\missing{i, s}\missing{j, s}(\measurement{i, s} - \measurement{j, s})^2}{\sum_{s \neq t} \missing{i, s}\missing{j, s}} \qtext{and} \distance{t, s}^{\col} := \frac{\sum_{j \neq i}\missing{j, t}\missing{j, s}(\measurement{j, t} - \measurement{j, s})^2}{\sum_{j \neq i} \missing{j, t}\missing{j, s}}
    \end{align}
    \item[\textbf{Step 2:}] (Construct neighborhood) Construct a row-wise and column-wise neighborhood of radius $\tuning_1$ and $\tuning_2$ respectively,
    \begin{align}
        \neighbors{t, \tuning_1}^\row := \big\{ j \in [N] \setminus \{i\} : \distance{i, j}^\row \leq \tuning \big\} \qtext{and} \neighbors{i, \tuning_2}^\col := \big\{ s \in [T] \setminus \{t\} : \distance{t, s}^\col \leq \tuning \big\}
    \end{align}
    \item[\textbf{Step 3:}] (Average across observed neighbors) Take the average of measurements within the neighborhood; the first and the second averaging correspond to $\drnn$ and $\tsnn$ respectively:
    \begin{align}\label{eq:drnn-tsnn-average}
        &\what{\theta}_{i, t, \eta}^{\mathsf{DR}} := \frac{\sum_{j \in \neighbors{t,\tuning_1}^\row, s \in \neighbors{i, \tuning_2}^\col}\missing{j, t}\missing{i, s}\missing{j, s}\big(\measurement{j, t} + \measurement{i, s} - \measurement{j, s}\big) }{\sum_{j \in \neighbors{t,\tuning_1}^\row, s \in \neighbors{i, \tuning_2}^\col}\missing{j, t}\missing{i, s}\missing{j, s}}\qtext{and} \\
        &\what{\theta}_{i, t, \eta}^{\mathsf{TS}} := \frac{\sum_{j \in \neighbors{t,\tuning_1}^\row, s \in \neighbors{i, \tuning_2}^\col}\missing{j, s}\measurement{j, s}}{\sum_{j \in \neighbors{t,\tuning_1}^\row, s \in \neighbors{i, \tuning_2}^\col}\missing{j, s}}.
    \end{align}
\end{itemize}

Next, we specity the exact implemention of the two modules $\distancemod$ and $\averagemod$ to recover $\tsnn$ and $\drnn$:

\begin{algorithm2e}[H]
\caption{$\tsnn$ for scalar matrix completion} 
  \label{algo:tsnn}
  \SetAlgoLined
  \DontPrintSemicolon
  \small
  {
\KwIn{\textup{$\mc Z, \mc A, \tuning,$ $(i, t)$}}
  \BlankLine
    Initialize entry-wise metric $\entrydistest{\measurement{j, s}, \measurement{j', s'}}\gets ( \measurement{j, s} - \measurement{j', s'} )^2$ and metric $\entrydist{x, y} \gets (x - y)^2$\\
    Initialize tuning parameter ${\tuning} \gets (\tuning_1, \tuning_2)$\\
    Calculate row-wise and column-wise metric $\big\{ \distance{i, j}^\row: j \neq i \big\}, \big\{ \distance{t, s}^\col: s \neq t \big\} \gets $ $\distancemod(\what \varphi, \measurement{}, \missing{})$ \\
    Initialize weight $\weight{j, s} \gets  \indic{\distance{i, j}^\row \leq \tuning_1, \distance{s, t}^\col \leq \tuning_2}$ \\
    Calculate average $\what \theta_{i, t} \gets \averagemod(\varphi, \mc W, \mc Z, \mc A)$
    \BlankLine
\KwRet{$\what \theta_{i, t}$}
}
\end{algorithm2e} 

For $\drnn$ algorithm below, we consider $\mc Z$ and $\mc A$ to be $N\times T$ sized matrices, so that their transpose is well defined. Then note that $\colnn$ is simply applying \cref{algo:vanilla-nn} with transposed observation matrices. 
\begin{algorithm2e}[H]
\caption{$\drnn$ for scalar matrix completion} 
  \label{algo:dsnn}
  \SetAlgoLined
  \DontPrintSemicolon
  \small
  {
\KwIn{\textup{$\mc Z, \mc A, \tuning,$ $(i, t)$}}
  \BlankLine
    Initialize $\rownn \gets$ \cref{algo:vanilla-nn} with inputs $(\mc Z, \mc A, \tuning, (i, t))$ and $\tuning \gets (\tuning_1, 0)$ \\
    Initialize $\colnn \gets$ \cref{algo:vanilla-nn} with input $(\mc Z^T, \mc A^T, \tuning, (i, t))$ and $\tuning \gets (\tuning_1, 0)$\\
    Initialize $\tsnn \gets$ \cref{algo:tsnn} with inputs $(\mc Z, \mc A, \tuning, (i, t))$ and ${\tuning} \gets (\tuning_1, \tuning_2)$ \\
    Calculate $\what \theta_{i, t} \gets \rownn + \colnn-\tsnn$
    \BlankLine
\KwRet{$\what \theta_{i, t}$}
}
\end{algorithm2e} 


\subsection{Distributional nearest neighbors}



Unlike the scalar nearest neighbor methods, distributional nearest neighbors necessitate a distributional notion of distance between rows and columns of matrix and a distributional analog of averaging. \cite{choi2024learning} and \cite{feitelberg2024distributional} use maximum mean discrepency~(in short $\mmd$) of kernel mean embeddings~\cite{muandet2017kernel} and Wasserstein metric~(in short $\wasserstein$)~\cite{bigot2020} respectively both for defining the distance between rows / columns and for averaging. The corresponding barycenters of $\mmd$ and $\wasserstein$~\cite{cohen2020estimating,bigot2017geodesic} are used for averaging, and so the methods are coined kernel nearest neighbors~(in short $\kernelnn$) and 
Wasserstein nearest neighbors~(in short $\wassersteinnn$) respectively.

We elaborate on a vanilla version three step procedure of $\kernelnn, \wassersteinnn$ that explicitly constructs neighborhoods. The input are measurements $\mc Z$, missingness $\mc A$, the target index $(i, t)$ and the radius $\tuning$,
\begin{itemize}
    \item[\textbf{Step 1:}] (Distance between rows) Calculate the distance between row $i$ and any row $j \in [N] \setminus \{i\}$ by averaging the estimator of distribution metric $\what \varrho$:
    \begin{align}\label{eq:snn-distance}
        \distance{i, j}^{\mmd} := \frac{\sum_{s \neq t}\missing{i, s}\missing{j, s} \what{\mmd}_\kernel^2 ( \measurement{i, s}, \measurement{j, s} ) }{\sum_{s \neq t} \missing{i, s}\missing{j, s}} \qtext{and} \distance{i, j}^{\wasserstein} := \frac{\sum_{s \neq t}\missing{i, s}\missing{j, s} \hatwasserstein^2 ( \measurement{i, s}, \measurement{j, s} ) }{\sum_{s \neq t} \missing{i, s}\missing{j, s}}.
    \end{align}
    \item[\textbf{Step 2:}] (Construct neighborhood) Construct a neighborhood of radius $\tuning$ within the $t$th column using the distances $\{\distance{i, j}:j \neq i\}$: 
    \begin{align}
        \neighbors{t, \tuning}^\mmd := \big\{ j \in [N] \setminus \{i\} : \distance{i, j}^\mmd \leq \tuning \big\} \qtext{and} \neighbors{t, \tuning}^{\wasserstein} := \big\{ j \in [N] \setminus \{i\} : \distance{i, j}^{\wasserstein} \leq \tuning \big\}
    \end{align}
    \item[\textbf{Step 3:}] (Average across observed neighbors) Set $\mu^Z_{i, t} = n^{-1}\sum_{\ell = 1}^{n} \dirac_{\sample{\ell}(i, t)}$ as the empirical measure of the multiple measurements $\measurement{i, t}$. Take the barycenter within the neighborhood:
    \begin{align}\label{eq:dnn-average}
        \what{\mu}^{\mmd}_{i, t, \tuning} &:= \frac{1}{|\neighbors{t, \tuning}^\mmd|} \sum_{j \in \neighbors{t, \tuning}^\mmd} \missing{j, t} \mu_{j, t}^Z \quad\text{and}\\
        \what{\mu}^{\wasserstein}_{i, t, \tuning} &:= \argmin_{\mu} \sum_{j \in \in \neighbors{t, \tuning}^{\wasserstein}} \wasserstein^2(\mu, \mu_{j, t}^Z).
    \end{align}
\end{itemize}
For further details on the $\wasserstein$ and $\mmd$ algorithms see \cite{feitelberg2024distributional} and \cite{choi2024learning}, respectively.

\begin{algorithm2e}[H]
\caption{Vanilla~(row-wise) distributional nearest neighbor} 
  \label{algo:vanilla-dnn}
  \SetAlgoLined
  \DontPrintSemicolon
  \small
  {
\KwIn{\textup{$\mc Z, \mc A, \kernel, \tuning,$ $(i, t)$}}
  \BlankLine
    Initialize entry-wise metric $\entrydistest{\measurement{j, s}, \measurement{j', s'}} \gets$ $\what{\mmd}^2_\kernel(\measurement{j, s}, \measurement{j', s'})$ or $\hatwasserstein^2(\measurement{j, s}, \measurement{j', s'})$ \\
    Initialize metric $\entrydist{x, y} \gets$ ${\mmd}^2_\kernel(x, y)$ or $\wasserstein^2(x, y)$ \\
    Initialize tuning parameter ${\tuning} \gets (\tuning_1, 0)$\\
    Calculate row-wise metric $\big\{ \distance{i, j}^\row: j \neq i \big\} \gets $ $\distancemod(\what\varphi, \measurement{}, \missing{})$ \\
    Initialize weight $\weight{j, s} \gets  \indic{\distance{i, j}^\row \leq \tuning_1, \distance{s, t}^\col \leq \tuning_2}$ \\
    Calculate average $\what \mu_{i, t} \gets \averagemod(\varphi, \mc W, \mc Z, \mc A)$
    \BlankLine
\KwRet{$\what \mu_{i, t}$}
}
\end{algorithm2e}

\subsection{Adaptively weighted nearest neighbors}
\label{app:star-nn}

We elaborate on the adaptive variant of the vanilla nearest neighbor algorithm $\awnn$ as mentioned in \cref{subsec: existing methods} and \cref{tab:summary}. The input are measurements $\mc Z$, and missingness $\mc A$. Note that there is no need for radius parameter $\eta$ and hence no CV.

\begin{itemize}
    \item[\textbf{Step 1:}] (Distance between rows and initial noise variance estimate) Calculate an estimate for noise variance and then the distance between any pair of distinct rows $i, j \in [N]$ by averaging the squared Euclidean distance across overlapping columns:
    \begin{align}\label{eq:snn-distance}
        &\distance{i, j} := \frac{\sum_{s \neq t}\missing{i, s}\missing{j, s}(\measurement{i, s} - \measurement{j, s})^2}{\sum_{s \neq t} \missing{i, s}\missing{j, s}},\qquad \overline{Z}\gets \frac{\sum_{j,s\in[N]\times[T]}A_{j,s}Z_{j,s}}{\sum_{j,s\in[N]\times[T]}A_{j,s}}, \\
        &\what{\sigma}^2\gets \frac{\sum_{j,s\in[N]\times[T]}A_{j,s}(Z_{j,s} - \overline{Z})^2}{\sum_{j,s\in[N]\times[T]}A_{j,s}}
    \end{align}
    
    \item[\textbf{Step 2:}] (Construct weights) For all rows and columns $(i,t)\in[N]\times[T]$, evaluate $ w^{(i, t)} = ( w_{1, t}, \cdots,  w_{n, t})$, the weights that optimally minimizes the following loss involving an estimate of the noise variance $\what\sigma^2$,
    \begin{align}
    \label{eq: awnn weights}
         w^{(i, t)}    =\mbox{arg min}_{\widehat w^{(i, t)}}\brackets{2\log(2m/\delta)\widehat \sigma^2\|\widehat w^{(i,t)} \|_2^2 + \sum_{i' \in [N]} \widehat w_{i', t} A_{i',t} \widehat \rho_{i', i} }, 
    \end{align}
where $\widehat w^{(i, t)} = (\widehat w_{1, t}, \cdots, \widehat w_{n, t})$ is a non-negative vector that satisfy $\sum_{i' = 1}^n \what w_{i', t}A_{i',t} = 1$.
    \item[\textbf{Step 3:}] (Weighted average) Take the weighted average of measurements:
    \begin{align}\label{eq:awnn-average}
        \widehat \theta_{i,t} = \sum_{i' \in[N]}\widehat w_{i', t} A_{i',t}X_{i',t}, \qquad \forall (i,t)\in[N]\times[T]
    \end{align}

    \item[\textbf{Step 4:}] (Fixed point iteration over noise variance) Obtain new estimate of noise variance and stop if difference between old and new $\what\sigma^2$ is small.
    \begin{align}
        \what{\sigma}^2\gets \frac{1}{\sum_{i\in[N],t\in[T]}\missing{i,t}}\sum_{i\in[N],t\in[T]}\parenth{Z_{i,t}-\what{\theta}_{i,t}}^2\missing{i,t}
    \end{align}
\end{itemize}

\paragraph{No cross-validation in AWNN} The optimization problem in \cref{eq: awnn weights} can be solved exactly in linear time (worst case complexity) using convex optimization~\cite{sadhukhan2025adaptivelyweightednearestneighborsmatrix}. $\awnn$ doesn't rely on radius parameter $\tuning$. Not only it automatically assigns neighbors to ${(i, t)^{th}}$ entry during its weight calculation(non-neighbors get zero weight), but also takes into account the distance of the neighbors from the ${(i, t)^{th}}$ entry. The closer neighors get higher weights and vice - versa.

We further specify the exact implementation of the two modules $\distancemod$, $\averagemod$ to recover $\awnn$:

\begin{algorithm2e}[H]
\caption{$\awnn$ for scalar nearest neighbor} 
  \label{algo:awnn}
  \SetAlgoLined
  \DontPrintSemicolon
  \small
  {
\KwIn{\textup{$\mc Z, \mc A, $ $(i, t)$}}
  \BlankLine
    Initialize entry-wise metric $\entrydistest{\measurement{j, s}, \measurement{j', s'}}\gets ( \measurement{j, s} - \measurement{j', s'} )^2$ and metric $\entrydist{x, y} \gets (x - y)^2$\\
    Initialize noise - variance estimate ${\sigma_\eps^2} \gets \mrm{Variance}\parenth{\braces{\measurement{i,t}}_{(i,t)\in[N]\times[T]}}$\\
    Calculate row-wise metric $\big\{ \distance{i, j}^\row: j \neq i \big\} \gets $ $\distancemod(\what{\varphi}, \mc Z, \mc A)$ \\
    Initialize weight $\braces{\weight{1, t},\dots, \weight{n, t}  }\gets\mbox{arg min}_{\widehat w^{(i, t)}}\brackets{2\log(2m/\delta)\widehat \sigma^2\|\widehat w^{(i,t)} \|_2^2 + \sum_{i' \in [N]} \widehat w_{i', t} A_{i',t} \widehat \rho_{i', i} }$ \\
    Calculate average $\what \theta_{i, t} \gets \averagemod(\varphi, \mc W, \measurement{}, \missing{})$
    \BlankLine
\KwRet{$\what \theta_{i, t}$}
}
\end{algorithm2e}

%% file: appendices/cv.tex
\section{Cross-Validation}\label{app:cv}

For each nearest neighbor method, we use cross-validation to optimize hyperparameters including distance thresholds and weights, depending on which nearest neighbor algorithm is chosen. Specifically, for each experiment, we choose a subset of the training test to optimize hyperparameters by masking those matrix cells and then estimating the masked values. We utilize the HyperOpt library \cite{bergstra2013making} to optimize (possibly multiple) hyperparamters using the Tree of Parzen Estimator \cite{bergstra2011algorithms}, a Bayesian optimization method. Our package supports both regular distance thresholds and percentile-based thresholds, which adapt to the distances calculated within the specific dataset.

%% file: appendices/case_study_details.tex
\section{Case Study Details}\label{app:case-studies}

The boxplots are generated using matplotlib's \cite{matplotlib} standard boxplot function. The box shows the first, second, and third quartiles. The bottom line shows the first quartile minus the 1.5$\times$ the interquartile range. The top line shows the third quartile plus 1.5$\times$ the interquartile range. All experiments are run on standard computing hardware (MacBook Pro with an M2 Pro CPU with 32 GB of RAM).

\subsection{Synthetic data generation}\label{app:synthetic_gen}

Generate $\measurement{i, t} = \sample{i, t} \sim N(\theta_{i, t}, \sigma^2)$, i.e., scalar matrix completion setting, with a linear factor structure $\theta_{i, t} = u_i v_t$. Row latent factors $u_i \in \real^4$ are i.i.d. generated across $i = 1, ..., N$, where each entry of $u_i$ follow a uniform distribution with support $[-0.5, 0.5]$; column latent factors $v_t \in \real^4$ are generated in an identical manner. The missingness is MCAR with propensity $p_{i, t} = 0.5$ for all $i$ and $t$. Further, the size of column and rows are identical $N = T$. For the left panel in \cref{fig:auto_perf}, the noise level is set as $\sigma = 0.001$ and for the right panel $\sigma = 1$.


\subsection{HeartSteps V1}

\begin{figure}
    \centering
    \includegraphics[width=1\linewidth]{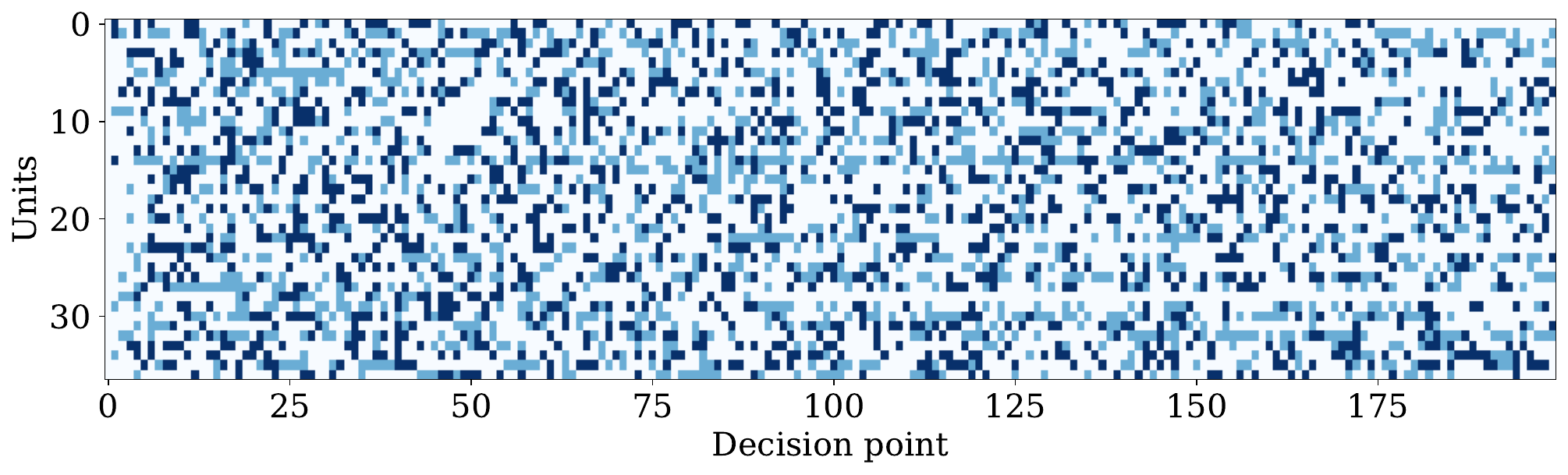}
    \caption{\tbf{HeartSteps V1 data notification pattern.} The dark blue entries indicate that the app sent a notification to a sedentary participant---the entry has value $A_{i, t} = 1$. The white entries indicate that the participant was available but did not receive a notification or they were active immediately prior to the decision point. The light blue entries indicate the participant was unavailable. We assign the value $A_{i, t} = 0$ for all the white and light blue entries.}
    \label{fig:heart_missing}
\end{figure}

The mobile application was designed to send notifications to users at various times during the day to encourage anti-sedentary activity such as stretching or walking. Participants could be marked as unavailable during decision points if they were in transit or snoozed their notifications, so notifications were only sent randomly if a participant was available and were never sent if they were unavailable. 
To process the data in the framework of \cref{eq:model}, we let matrix entry $Z_{i, t}$ be the average one hour step count for participant $i$ and decision point $t$ when a notification is sent (i.e. $A_{i, t} = 1$) and unknown when a notification is not sent (i.e. $A_{i, t} = 0$). The treatment assignment pattern is represented as the 37 x 200 matrix visualized in \cref{fig:heart_missing}.  
We use the dataset downloaded from \url{https://github.com/klasnja/HeartStepsV1} (CC-BY-4.0 License).

\subsection{MovieLens}\label{app:movielens}
We load MovieLens via a custom \texttt{MovieLensDataLoader} that (i) downloads and caches the \texttt{ml‑1m.zip} archive, (ii) reads \texttt{ratings.dat} into a user $\times$ movie pivot table, and (iii) constructs the binary mask where observed entries correspond to rated user–movie pairs. The data matrix is $\measurement{} \in\{1,\dots,5\}^{6040\times3952}$ and mask matrix is $\missing{} \in\{0,1\}^{6040\times3952}$. The data can be downloaded from \url{https://grouplens.org/datasets/movielens/1m/}. See \url{https://files.grouplens.org/datasets/movielens/ml-1m-README.txt} for the usage license.

\subsection{Proposition 99}
\label{app:prop99}

Data comes primarily from the Tax Burden on Tobacco compiled by Orzechowski and Walker~\cite{cdcBurdenTobacco} (ODC-By License). Using synthetic control methods, Abadie et al. construct a weighted combination of control states that closely resembles California's pre-1988 characteristics and cigarette consumption patterns. The optimal weights produce a synthetic California primarily composed of Colorado (0.164), Connecticut (0.069), Montana (0.199), Nevada (0.234), and Utah (0.334), with all other states receiving zero weight. The treatment effect is estimated as the difference between actual California per-capita cigarette sales and those of synthetic California after Proposition 99's implementation. By 2000, this analysis revealed that annual per-capita cigarette sales in California were approximately 26 packs lower than what they would have been without Proposition 99, representing about a 25\% reduction in cigarette consumption. To validate these findings, the authors conducted placebo tests by applying the same methodology to states not implementing tobacco control programs, confirming that California's reduction was unusually large and statistically significant (p = 0.026).

Proposition 99, the California Tobacco Tax and Health Protection Act of 1988, dataset spans from 1970 to 2000, providing 19 years of pre-intervention data before Proposition 99 was implemented in 1988 and 12 years of post-intervention data. It provides annual state-level cigarette consumption measured as per capita cigarette sales in packs based on tax revenue data. This data serves as a real data benchmark for many of the variants of synthetic controls~\cite{athey2021matrix}. We use the CDC dataset for the Nearest Neighbors methods and only use the target variable (i.e., cigarette consumption measured in packs per capita), and the dataset from SyntheticControlMethods library\footnote{\url{https://github.com/OscarEngelbrektson/SyntheticControlMethods/tree/master} (Apache-2.0 License)} for the SC baseline, since it relies on additional covariates.

\subsection{PromptEval}

MMLU is a multiple choice Q\&A benchmark with $57$ tasks, with a total of near $14$K examples\footnote{\url{https://github.com/felipemaiapolo/prompteval} (MIT License)}. $15$ different models, e.g., variants of Llama 3~\cite{llama3}, Mistral~\cite{jiang2023mistral} and Gemma~\cite{team2024gemma}. The examples are fed into the models with $100$ different varying prompting templates. The prompt templates are created by traversing between $3$ node modules, namely a \textit{separator}, a \textit{space} and an \textit{operator}~(see \cite[Alg. 3, App. J]{polo2024efficient} for details), from which $100$ unique prompt templates are created. The unitxt~\cite{bandel2024unitxt} preprocessing library is used to construct the dataset and evaluation is done by LM-Eval-Harness~\cite{eval-harness} library. The number of examples differ per task and each examples are evaluated on a model~(verifiable, so assigned $0$ or $1$ for correctness) by wrapping the examples with $100$ different prompt templates.

%% file: ref.bib
@misc{eval-harness,
  author       = {Gao, Leo and Tow, Jonathan and Abbasi, Baber and Biderman, Stella and Black, Sid and DiPofi, Anthony and Foster, Charles and Golding, Laurence and Hsu, Jeffrey and Le Noac'h, Alain and Li, Haonan and McDonell, Kyle and Muennighoff, Niklas and Ociepa, Chris and Phang, Jason and Reynolds, Laria and Schoelkopf, Hailey and Skowron, Aviya and Sutawika, Lintang and Tang, Eric and Thite, Anish and Wang, Ben and Wang, Kevin and Zou, Andy},
  title        = {A framework for few-shot language model evaluation},
  month        = 12,
  year         = 2023,
  publisher    = {Zenodo},
  version      = {v0.4.0},
  doi          = {10.5281/zenodo.10256836},
  url          = {https://zenodo.org/records/10256836}
}

@article{bandel2024unitxt,
  title={Unitxt: Flexible, shareable and reusable data preparation and evaluation for generative ai},
  author={Bandel, Elron and Perlitz, Yotam and Venezian, Elad and Friedman-Melamed, Roni and Arviv, Ofir and Orbach, Matan and Don-Yehyia, Shachar and Sheinwald, Dafna and Gera, Ariel and Choshen, Leshem and others},
  journal={arXiv preprint arXiv:2401.14019},
  year={2024}
}

@misc{llama3,
  author = {Meta},
  title = {Introducing Meta Llama 3: The most capable openly available LLM to date},
  year = {2024},
  howpublished = "\url{https://ai.meta.com/blog/meta-llama-3}"
}

@article{jiang2023mistral,
  title={Mistral 7B},
  author={Jiang, Albert Q and Sablayrolles, Alexandre and Mensch, Arthur and Bamford, Chris and Chaplot, Devendra Singh and Casas, Diego de las and Bressand, Florian and Lengyel, Gianna and Lample, Guillaume and Saulnier, Lucile and others},
  journal={arXiv preprint arXiv:2310.06825},
  year={2023}
}

@article{team2024gemma,
  title={Gemma: Open models based on gemini research and technology},
  author={Team, Gemma and Mesnard, Thomas and Hardin, Cassidy and Dadashi, Robert and Bhupatiraju, Surya and Pathak, Shreya and Sifre, Laurent and Rivi{\`e}re, Morgane and Kale, Mihir Sanjay and Love, Juliette and others},
  journal={arXiv preprint arXiv:2403.08295},
  year={2024}
}

@article{bigot2020,
	author = {{Bigot, J{\'e}r{\'e}mie}},
	journal = {ESAIM: ProcS},
	pages = {1-19},
	title = {Statistical data analysis in the Wasserstein space*},
	volume = 68,
	year = 2020
}

@article{recht2011simpler,
  title={A simpler approach to matrix completion.},
  author={Recht, Benjamin},
  journal={Journal of Machine Learning Research},
  volume={12},
  number={12},
  year={2011}
}

@article{klasnja2019efficacy,
  title={Efficacy of contextually tailored suggestions for physical activity: a micro-randomized optimization trial of HeartSteps},
  author={Klasnja, Predrag and Smith, Shawna and Seewald, Nicholas J and Lee, Andy and Hall, Kelly and Luers, Brook and Hekler, Eric B and Murphy, Susan A},
  journal={Annals of Behavioral Medicine},
  volume={53},
  number={6},
  pages={573--582},
  year={2019},
  publisher={Oxford University Press US}
}

@article{athey2022design,
  title={Design-based analysis in difference-in-differences settings with staggered adoption},
  author={Athey, Susan and Imbens, Guido W},
  journal={Journal of Econometrics},
  volume={226},
  number={1},
  pages={62--79},
  year={2022},
  publisher={Elsevier}
}

@article{cannings2020local,
  title={Local nearest neighbour classification with applications to semi-supervised learning},
  author={Cannings, Timothy I and Berrett, Thomas B and Samworth, Richard J},
  journal={The Annals of Statistics},
  volume={48},
  number={3},
  pages={1789--1814},
  year={2020},
  publisher={JSTOR}
}

@article{sadhukhan2025adaptivelyweightednearestneighborsmatrix,
      title={Adaptively-weighted Nearest Neighbors for Matrix Completion}, 
      author={Tathagata Sadhukhan and Manit Paul and Raaz Dwivedi},
         journal={arXiv preprint arXiv:2505.09612},
       year={2025},
        url={https://arxiv.org/abs/2505.09612}
}

@misc{cdcBurdenTobacco,
	author = {Orzechowski and Walker},
	title = {{T}he {T}ax {B}urden on {T}obacco, 1970-2019 | {D}ata | {C}enters for {D}isease {C}ontrol and {P}revention --- data.cdc.gov},
	howpublished = {\url{https://data.cdc.gov/api/views/7nwe-3aj9/rows.csv?accessType=DOWNLOAD}},
	year = {2023},
	note = {[Accessed 16-05-2025]},
}

@article{cover1967nearest,
  title={Nearest neighbor pattern classification},
  author={Cover, Thomas and Hart, Peter},
  journal={IEEE transactions on information theory},
  volume={13},
  number={1},
  pages={21--27},
  year={1967},
  publisher={IEEE}
}

@book{little2019statistical,
  title={Statistical analysis with missing data},
  author={Little, Roderick JA and Rubin, Donald B},
  volume={793},
  year={2019},
  publisher={John Wiley \& Sons}
}

@article{athey2021matrix,
  title={Matrix completion methods for causal panel data models},
  author={Athey, Susan and Bayati, Mohsen and Doudchenko, Nikolay and Imbens, Guido and Khosravi, Khashayar},
  journal={Journal of the American Statistical Association},
  volume={116},
  number={536},
  pages={1716--1730},
  year={2021},
  publisher={Taylor \& Francis}
}

@article{candes2012exact,
  title={Exact matrix completion via convex optimization},
  author={Candes, Emmanuel and Recht, Benjamin},
  journal={Communications of the ACM},
  volume={55},
  number={6},
  pages={111--119},
  year={2012},
  publisher={ACM New York, NY, USA}
}

@article{scikit-learn,
  title={Scikit-learn: Machine Learning in {P}ython},
  author={Pedregosa, F. and Varoquaux, G. and Gramfort, A. and Michel, V.
          and Thirion, B. and Grisel, O. and Blondel, M. and Prettenhofer, P.
          and Weiss, R. and Dubourg, V. and Vanderplas, J. and Passos, A. and
          Cournapeau, D. and Brucher, M. and Perrot, M. and Duchesnay, E.},
  journal={Journal of Machine Learning Research},
  volume={12},
  pages={2825--2830},
  year={2011}
}

@article{dwivedi2022personalization,
  title={Deep dive into personalization},
  author={Dwivedi, Raaz and Zhang, Kelly and Chhabaria, Prasidh and Murphy, Susan},
  journal={Working paper},
  year={2022},
}

@article{muandet2017kernel,
  title={Kernel mean embedding of distributions: A review and beyond},
  author={Muandet, Krikamol and Fukumizu, Kenji and Sriperumbudur, Bharath and Sch{\"o}lkopf, Bernhard and others},
  journal={Foundations and Trends{\textregistered} in Machine Learning},
  volume={10},
  number={1-2},
  pages={1--141},
  year={2017},
  publisher={Now Publishers, Inc.}
}

@article{bhattacharya2022matrix,
  title={Matrix completion with data-dependent missingness probabilities},
  author={Bhattacharya, Sohom and Chatterjee, Sourav},
  journal={IEEE Transactions on Information Theory},
  volume={68},
  number={10},
  pages={6762--6773},
  year={2022},
  publisher={IEEE}
}

@article{abadie2010synthetic,
  title={Synthetic control methods for comparative case studies: Estimating the effect of California’s tobacco control program},
  author={Abadie, Alberto and Diamond, Alexis and Hainmueller, Jens},
  journal={Journal of the American statistical Association},
  volume={105},
  number={490},
  pages={493--505},
  year={2010},
  publisher={Taylor \& Francis}
}

@article{chatterjee2015matrix,
	author = {Sourav Chatterjee},
	doi = {10.1214/14-AOS1272},
	journal = {The Annals of Statistics},
	number = {1},
	pages = {177 -- 214},
	publisher = {Institute of Mathematical Statistics},
	title = {{Matrix estimation by Universal Singular Value Thresholding}},
	url = {https://doi.org/10.1214/14-AOS1272},
	volume = {43},
	year = {2015}
}

@inproceedings{agarwal2023causal,
  title={Causal matrix completion},
  author={Agarwal, Anish and Dahleh, Munther and Shah, Devavrat and Shen, Dennis},
  booktitle={The Thirty Sixth Annual Conference on Learning Theory},
  pages={3821--3826},
  year={2023},
  organization={PMLR}
}

@article{dwivedi2022doubly,
  title={Doubly robust nearest neighbors in factor models},
  author={Dwivedi, Raaz and Tian, Katherine and Tomkins, Sabina and Klasnja, Predrag and Murphy, Susan and Shah, Devavrat},
  journal={arXiv preprint arXiv:2211.14297},
  year={2022}
}

@article{li2019nearest,
  title={Nearest neighbors for matrix estimation interpreted as blind regression for latent variable model},
  author={Li, Yihua and Shah, Devavrat and Song, Dogyoon and Yu, Christina Lee},
  journal={IEEE Transactions on Information Theory},
  volume={66},
  number={3},
  pages={1760--1784},
  year={2019},
  publisher={IEEE}
}

@article{cohen2020estimating,
  title={Estimating barycenters of measures in high dimensions},
  author={Cohen, Samuel and Arbel, Michael and Deisenroth, Marc Peter},
  journal={arXiv preprint arXiv:2007.07105},
  year={2020}
}

@article{bigot2017geodesic,
	author = {J{\'e}r{\'e}mie Bigot and Ra{\'u}l Gouet and Thierry Klein and Alfredo L{\'o}pez},
	journal = {Annales de l'Institut Henri Poincar{\'e}, Probabilit{\'e}s et Statistiques},
	number = {1},
	pages = {1 -- 26},
	title = {{Geodesic PCA in the Wasserstein space by convex PCA}},
	volume = {53},
	year = {2017}
}

@article{dwivedi2022counterfactual,
  title={Counterfactual inference for sequential experiments},
  author={Dwivedi, Raaz and Tian, Katherine and Tomkins, Sabina and Klasnja, Predrag and Murphy, Susan and Shah, Devavrat},
  journal={arXiv preprint arXiv:2202.06891},
  year={2022}
}

@article{bai2009panel,
  title={Panel data models with interactive fixed effects},
  author={Bai, Jushan},
  journal={Econometrica},
  volume={77},
  number={4},
  pages={1229--1279},
  year={2009},
  publisher={Wiley Online Library}
}

@article{hastie2015matrix,
  title={Matrix completion and low-rank SVD via fast alternating least squares},
  author={Hastie, Trevor and Mazumder, Rahul and Lee, Jason D and Zadeh, Reza},
  journal={The Journal of Machine Learning Research},
  volume={16},
  number={1},
  pages={3367--3402},
  year={2015},
  publisher={JMLR. org}
}

@article{ma2019missing,
  title={Missing not at random in matrix completion: The effectiveness of estimating missingness probabilities under a low nuclear norm assumption},
  author={Ma, Wei and Chen, George H},
  journal={Advances in neural information processing systems},
  volume={32},
  year={2019}
}

@article{feitelberg2024distributional,
  title={Distributional Matrix Completion via Nearest Neighbors in the Wasserstein Space},
  author={Feitelberg, Jacob and Choi, Kyuseong and Agarwal, Anish and Dwivedi, Raaz},
  journal={arXiv preprint arXiv:2410.13112},
  year={2024}
}

@article{choi2024learning,
  title={Learning Counterfactual Distributions via Kernel Nearest Neighbors},
  author={Choi, Kyuseong and Feitelberg, Jacob and Chin, Caleb and Agarwal, Anish and Dwivedi, Raaz},
  journal={arXiv preprint arXiv:2410.13381},
  year={2024}
}

@article{hendrycks2020measuring,
  title={Measuring massive multitask language understanding},
  author={Hendrycks, Dan and Burns, Collin and Basart, Steven and Zou, Andy and Mazeika, Mantas and Song, Dawn and Steinhardt, Jacob},
  journal={arXiv preprint arXiv:2009.03300},
  year={2020}
}

@article{sadhukhan2024adaptivity,
  title={On adaptivity and minimax optimality of two-sided nearest neighbors},
  author={Sadhukhan, Tathagata and Paul, Manit and Dwivedi, Raaz},
  journal={arXiv preprint arXiv:2411.12965},
  year={2024}
}

@inproceedings{polo2024efficient,
  title={Efficient multi-prompt evaluation of LLMs},
  author={Polo, Felipe Maia and Xu, Ronald and Weber, Lucas and Silva, M{\'\i}rian and Bhardwaj, Onkar and Choshen, Leshem and de Oliveira, Allysson Flavio Melo and Sun, Yuekai and Yurochkin, Mikhail},
  booktitle={The Thirty-eighth Annual Conference on Neural Information Processing Systems},
  year={2024}
}

@inproceedings{sarwar2001item,
    author = {Sarwar, Badrul and Karypis, George and Konstan, Joseph and Riedl, John},
    title = {Item-based collaborative filtering recommendation algorithms},
    year = {2001},
    isbn = {1581133480},
    publisher = {Association for Computing Machinery},
    address = {New York, NY, USA},
    url = {https://doi.org/10.1145/371920.372071},
    doi = {10.1145/371920.372071},
    booktitle = {Proceedings of the 10th International Conference on World Wide Web},
    pages = {285–295},
    numpages = {11},
    location = {Hong Kong, Hong Kong},
    series = {WWW '01}
}

@article{koren2009matrix,
  author={Koren, Yehuda and Bell, Robert and Volinsky, Chris},
  journal={Computer}, 
  title={Matrix Factorization Techniques for Recommender Systems}, 
  year={2009},
  volume={42},
  number={8},
  pages={30-37},
  doi={10.1109/MC.2009.263}
}

@article{keshavan2010matrix,
  title={Matrix completion from a few entries},
  author={Keshavan, Raghunandan H and Montanari, Andrea and Oh, Sewoong},
  journal={IEEE transactions on information theory},
  volume={56},
  number={6},
  pages={2980--2998},
  year={2010},
  publisher={IEEE}
}

@article{movielens,
author = {Harper, F. Maxwell and Konstan, Joseph A.},
title = {The MovieLens Datasets: History and Context},
year = {2015},
issue_date = {January 2016},
publisher = {Association for Computing Machinery},
address = {New York, NY, USA},
volume = {5},
number = {4},
issn = {2160-6455},
url = {https://doi.org/10.1145/2827872},
doi = {10.1145/2827872},
journal = {ACM Trans. Interact. Intell. Syst.},
month = dec,
articleno = {19},
numpages = {19}
}

@inproceedings{bergstra2013making,
  title={Making a science of model search: Hyperparameter optimization in hundreds of dimensions for vision architectures},
  author={Bergstra, James and Yamins, Daniel and Cox, David},
  booktitle={International conference on machine learning},
  pages={115--123},
  year={2013},
  organization={PMLR}
}

@article{bergstra2011algorithms,
  title={Algorithms for hyper-parameter optimization},
  author={Bergstra, James and Bardenet, R{\'e}mi and Bengio, Yoshua and K{\'e}gl, Bal{\'a}zs},
  journal={Advances in neural information processing systems},
  volume={24},
  year={2011}
}

@Article{matplotlib,
  Author    = {Hunter, J. D.},
  Title     = {Matplotlib: A 2D graphics environment},
  Journal   = {Computing in Science \& Engineering},
  Volume    = {9},
  Number    = {3},
  Pages     = {90--95},
  publisher = {IEEE COMPUTER SOC},
  doi       = {10.1109/MCSE.2007.55},
  year      = 2007
}

@article{liao2019rlheartsteps,
	author = {Liao, Peng and Greenewald, Kristjan and Klasnja, Predrag and Murphy, Susan},
	journal = {Proc. ACM Interact. Mob. Wearable Ubiquitous Technol.},
	month = mar,
	number = {1},
	title = {Personalized HeartSteps: A Reinforcement Learning Algorithm for Optimizing Physical Activity},
	volume = {4},
	year = {2020}
}

@article{qian2022microrandomized,
  title={The microrandomized trial for developing digital interventions: Experimental design and data analysis considerations.},
  author={Qian, Tianchen and Walton, Ashley E and Collins, Linda M and Klasnja, Predrag and Lanza, Stephanie T and Nahum-Shani, Inbal and Rabbi, Mashfiqui and Russell, Michael A and Walton, Maureen A and Yoo, Hyesun and others},
  journal={Psychological methods},
  volume={27},
  number={5},
  pages={874},
  year={2022},
  publisher={American Psychological Association}
}

@inproceedings{gamma1993design,
  title={Design patterns: Abstraction and reuse of object-oriented design},
  author={Gamma, Erich and Helm, Richard and Johnson, Ralph and Vlissides, John},
  booktitle={ECOOP’93—Object-Oriented Programming: 7th European Conference Kaiserslautern, Germany, July 26--30, 1993 Proceedings 7},
  pages={406--431},
  year={1993},
  organization={Springer}
}

@article{geathers2025benchmarking,
  title={Benchmarking Generative AI for Scoring Medical Student Interviews in Objective Structured Clinical Examinations (OSCEs)},
  author={Geathers, Jadon and Hicke, Yann and Chan, Colleen and Rajashekar, Niroop and Sewell, Justin and Cornes, Susannah and Kizilcec, Rene and Shung, Dennis},
  journal={arXiv preprint arXiv:2501.13957},
  year={2025}
}
